\documentclass[lettersize,journal]{IEEEtran}
\usepackage{amsmath,amsfonts}
\usepackage{algorithmic}
\usepackage{algorithm}
\usepackage{array}
\usepackage[caption=false,font=normalsize,labelfont=sf,textfont=sf]{subfig}
\usepackage{textcomp}
\usepackage{stfloats}
\usepackage{url}
\usepackage{verbatim}
\usepackage{graphicx}
\usepackage{booktabs}
\usepackage{multirow}
\usepackage{cite}
\newcommand{\xhdr}[1]{{\noindent\bfseries #1}.}

\begin{document}

\title{StdGEN++: A Comprehensive System for Semantic-Decomposed 3D Character Generation}

\author{Yuze He, Yanning Zhou, Wang Zhao, Jingwen Ye, Zhongkai Wu, Ran Yi, Yong-Jin Liu,~\IEEEmembership{Senior Member,~IEEE}
\thanks{Y.~He, W.~Zhao and Y.-J.~Liu are with the Department of Computer Science and Technology, Tsinghua University, Beijing, China.}
\thanks{Y.~Zhou, J.~Ye, Z.~Wu are with Tencent AIPD, Shenzhen, China.}
\thanks{R.~Yi, is with School of Computer Science, Shanghai Jiao Tong University, Shanghai, China.}
\thanks{Y.~Zhou and Y.-J.~Liu are the corresponding authors. E-mail: ynzhou0907@gmail.com, liuyongjin@tsinghua.edu.cn}
}

\maketitle

\begin{abstract}
We present StdGEN++, a novel and comprehensive system for generating high-fidelity, semantically decomposed 3D characters from diverse inputs.
Existing 3D generative methods often produce monolithic meshes that lack the structural flexibility required by industrial pipelines in gaming and animation.
Addressing this gap, StdGEN++ is built upon a Dual-branch Semantic-aware Large Reconstruction Model (Dual-Branch S-LRM), which jointly reconstructs geometry, color, and per-component semantics in a feed-forward manner.
To achieve production-level fidelity, we introduce a novel semantic surface extraction formalism compatible with hybrid implicit fields. This mechanism is accelerated by a coarse-to-fine proposal scheme, which significantly reduces memory footprint and enables high-resolution mesh generation.
Furthermore, we propose a video-diffusion-based texture decomposition module that disentangles appearance into editable layers (e.g., separated iris and skin), resolving semantic confusion in facial regions.
Experiments demonstrate that StdGEN++ achieves state-of-the-art performance, significantly outperforming existing methods in geometric accuracy and semantic disentanglement.
Crucially, the resulting structural independence unlocks advanced downstream capabilities, including non-destructive editing, physics-compliant animation, and gaze tracking, making it a robust solution for automated character asset production.
\end{abstract}


\section{Introduction}
\label{sec:intro}

Generating high-quality 3D characters from single images has widespread applications in virtual reality, video games, filmmaking, etc. Beyond automatically creating a complete 3D character, there is an increasing demand for the ability to produce decomposable characters, where distinct semantic components like the body, clothes, and hair are disentangled. This decomposition allows for much easier editing, control, and animation of characters, greatly enhancing their usability across various downstream applications. 

However, creating such decomposable characters from single images is challenging, as each component may face issues such as occlusion, ambiguity, and inconsistencies in their interactions with other components. Existing methods for decomposable avatar generation primarily focus on realistic clothed human models, exploring disentangled 3D parametric~\cite{wang2023disentangled}, explicit~\cite{peng2024pica, pan2024humansplat}, or implicit~\cite{hong2022eva3d, huang2023avatarfusion, wang2024humancoser, dong2024tela} representations alongside various optimization techniques. These optimization approaches often employ score distillation loss~\cite{poole2022dreamfusiontextto3dusing2d} to leverage 2D generative priors, which leads to prolonged optimization times and the generation of coarse, high-contrast textures. Additionally, the dependence on parametric human models, such as SMPL-X~\cite{loper2023smpl}, is inadequate for virtual characters, which often exhibit exaggerated body proportions and complex clothing designs. 

CharacterGen~\cite{peng2024charactergen} was developed to efficiently generate characters from single images using a multi-view diffusion model and large reconstruction model~\cite{hong2023lrm} to address these limitations. Despite showing impressive generation capabilities in various posed images, 
CharacterGen can only produce holistic avatars in watertight meshes with no decomposability. These meshes require significant manual labor to separate, edit, or animate, limiting their applicability. Moreover, generated mesh quality is often unsatisfactory, particularly in finer details such as the character’s face and clothing, as shown in Fig.~\ref{fig:qual}. Therefore, efficiently generating high-quality, decomposable 3D characters remains an open challenge. 

To address the above challenges, previous work StdGEN~\cite{he2025stdgen} proposed an efficient pipeline for generating semantically decomposed, high-quality 3D characters from a single image.
StdGEN introduced a Semantic-aware Large Reconstruction Model (S-LRM) that extends the original LRM with semantic awareness, enabling feed-forward reconstruction of unified geometry, color, and per-part semantics.
It further employed a differentiable multi-layer surface extraction scheme, supported by a specialized multi-view diffusion model and iterative refinement.
While StdGEN achieved promising results in generating A-pose characters, its reconstructions still exhibit limitations in resolution constraint, local detail fidelity (e.g., facial features), input modality flexibility, and texture decomposability—all of which hinder its direct deployment in industrial pipelines.

In this paper, we substantially improve upon StdGEN~\cite{he2025stdgen} and propose \textbf{StdGEN++}, a comprehensive system for generating high-fidelity, semantically decomposed 3D characters with superior industrial compatibility.
Building upon the foundation of StdGEN, we introduce significant architectural upgrades and novel functional modules:

\begin{itemize}
    \item \textbf{\textit{Dual-branch Architecture and High-Resolution Extraction.}} 
    Generating industrial-grade characters requires precise control over both global structure and fine-grained details, which the single-branch model in StdGEN struggles to balance. To this end, we propose a \textbf{Dual-branch S-LRM}, enhanced with two specialized LoRA adapters: one for global body structure and another for fine-grained facial semantics.
    Furthermore, to overcome the resolution bottleneck, we upgrade the semantic surface extraction formalism (originally introduced in StdGEN) by integrating it with a novel coarse-to-fine proposal scheme. This mechanism efficiently reduces memory costs, enabling high-resolution output.
    Combined with a structure-aware hole-filling regularization, this design achieves substantially higher geometric accuracy and surface integrity compared to the single-branch baseline, effectively resolving critical artifacts like clothing tears and facial distortions.
    \item \textbf{\textit{Generative Texture Decomposition for Industrial Standards.}} 
    While standard production pipelines demand layered textures for editing and gaze tracking, StdGEN is restricted to monolithic atlases that fundamentally limit such downstream capabilities. We address this by designing a video-diffusion-based texture decomposition paradigm. 
    By formulating semantic layers as temporal frames, our model leverages spatial-temporal attention to not only disentangle components (e.g., iris, eyelash, skin) but also \textbf{generatively inpaint occluded regions} (e.g., restoring clean eye white behind the iris). 
    This module, new to StdGEN++, yields spatially distinct and editable layers, directly enabling downstream tasks like gaze tracking.
    \item \textbf{\textit{Unified Input System and Advanced Dataset.}} 
    While StdGEN primarily focused on image-based canonicalization, StdGEN++ elevates this mechanism into a universal input framework.
    We establish the canonical A-pose as a standardized interface that seamlessly bridges diverse modalities—from abstract text prompts to unconstrained reference images.
    Supporting this system, we substantially extend the Anime3D++ dataset to present \textbf{Anime3D-EX}. This comprehensive resource adds three key components to the original 10,811 characters: (1) rich textual captions for cross-modal conditioning; (2) multi-scale head-centric renderings; and (3) disentangled ground-truth facial texture layers (e.g., separated iris, skin, and lashes).
    These additions provide the essential data foundation for high-fidelity facial reconstruction, generative texture decomposition, and text-driven generation, establishing a robust benchmark for future research.
\end{itemize}

Extensive experiments demonstrate that StdGEN++ achieves state-of-the-art reconstruction quality. Its structural independence and system-level flexibility lead to a robust solution that effectively bridges the gap between AI generation and professional 3D production workflows.

\section{Related Works}
\label{sec:related}

\subsection{3D Generation}
To circumvent the need for extensive 3D assets during training, several approaches suggest lifting powerful 2D pre-trained diffusion models~\cite{dhariwal2021diffusion,nichol2021glide,rombach2022high,saharia2022photorealistic} for 3D generation.
The earliest works~\cite{poole2022dreamfusiontextto3dusing2d, wang2022scorejacobianchaininglifting} incorporate a pre-trained 2D diffusion model for probability density distillation using Score Distillation Sampling (SDS).
These approaches gradually optimize a randomly initialized radiance field~\cite{sun2022direct,chen2022tensorf,barron2022mip} with volume rendering, making it time-consuming to generate an object.
Later research continues to enhance the aesthetics and accuracy of 3D content generation~\cite{lin2023magic3d, chen2023fantasia3d, tsalicoglou2023textmesh, shen2021dmtet, wang2023prolificdreamer} and further investigate different application scenarios~\cite{haque2023instruct,shao2023control4d,singer2023text,raj2023dreambooth3d}.
However, relying solely on 2D priors for 3D generation often leads to poor geometry representation, e.g., multi-faced Janus problem, due to the challenges in controlling precise viewpoints through text prompts.
The large-scale 3D datasets, e.g. Objaverse~\cite{deitke2023objaversexluniverse10m3d}, unlock the possibility of imposing 3D priors to the model. 
Several works utilize view-consistent images to fine-tune the diffusion model.  
Zero-1-to-3~\cite{liu2023zero} integrates 3D priors into 2D stable diffusion by fine-tuning the pre-trained model for novel view synthesis (NVS). To further enhance the multi-view consistency, several recent works~\cite{shimvdream, long2024wonder3d, liu2023syncdreamer, huang2024epidiff} propose synchronously generating multi-view images in a single generation process and achieving constraints in 3D place through feature interaction in attention mechanism.
Besides, the 3D native generation method shows powerful geometric generation ability~\cite{zhang2024clay,li2024craftsman,lu2024direct2}. 
However, the ability of these methods to follow instructions is typically moderate; therefore, they face challenges in achieving the desired outcomes in scenarios requiring precise restoration of reference images, e.g., 3D character generation.

\subsection{Large Reconstruction Model}
Large Reconstruction Model~(LRM)~\cite{hong2023lrm} leverages the transformer-based model to map the single image feature to implicit tri-plane representation. 
Instant3D~\cite{liinstant3d} extends LRM by feeding multi-view images instead of a single image.
LGM~\cite{tang2024lgm}, GRM~\cite{xu2024grm} and GS-LRM~\cite{gslrm2024} replace the 3D representation to 3D Gaussians, embracing its efficiency in rendering and low memory consumption.
InstantMesh~\cite{xu2024instantmesh} and CRM~\cite{wang2024crm} explicitly model the geometry by equipping the generative pipeline with FlexiCubes~\cite{shen2023flexible}, achieving high-quality surface extraction and high rendering speed. 
The following works further explored applying advanced model architecture~\cite{zhang2024geolrmgeometryawarelargereconstruction} or 3D representation~\cite{chen2024lara, cui2024lam3d}, aiming to improve the efficiency, realism, and generalization of reconstruction.
Integrating with multi-view diffusion models, these LRMs can achieve text-to-3D generation or single image-to-3D generation.
Yet all these methods typically produce holistic models.
In contrast, our method generates semantically decomposed characters, making downstream processing such as editing and animation much more efficient.

\subsection{3D Character Generation}

3D character generation is a challenging problem due to its high precision requirements and the scarcity of data.
One line of work leverages 3D-aware GANs to model the distribution of digital humans
~\cite{bergman2023generativeneuralarticulatedradiance, hong2022eva3d, jiang2022humangengeneratinghumanradiance, zhang2022avatargen, noguchi2022unsupervisedlearningefficientgeometryaware}.
Recently the SDS-based methods have shown the possibility of generating a variety of stylized characters~\cite{cao2023dreamavatartextandshapeguided3d, huang2023dreamwaltzmakescenecomplex, wang2023disentangled, dong2024tela, kim2024gala}, yet it suffers from the long optimization times and the difficulty of meticulous style control.
Frankenstein~\cite{yan2024frankenstein} concentrates on producing decomposed, textureless 3D meshes based on 2D layouts, restricting the potential for achieving high-fidelity reconstruction from the reference image.
CharacterGen~\cite{peng2024charactergen} calibrates input poses to canonical multi-view images via an image-conditioned multi-view diffusion model, followed by LRM for 3D character reconstruction and multi-view texture back projection, but still exhibits limited geometry and texture quality.
Our approach, in contrast, employs a semantic-aware, feed-forward paradigm that generates high-quality, decomposable characters using only one forward pass from an arbitrary reference image, providing significant efficiency and quality improvement.

\section{Anime3D-EX Dataset}
\label{sec:dataset}
We introduce \textbf{Anime3D-EX}, a substantial extension of the Anime3D++ dataset~\cite{he2025stdgen}, tailored to support the advanced facial specialization, multimodal input, and texture decomposition features of StdGEN++.
Starting from an initial collection of $\sim$14k models from VRoid-Hub, we apply a rigorous cleaning pipeline to curate 10,811 high-quality 3D anime characters.

\begin{figure*}[tb]
\centering
\includegraphics[width=0.90\linewidth]{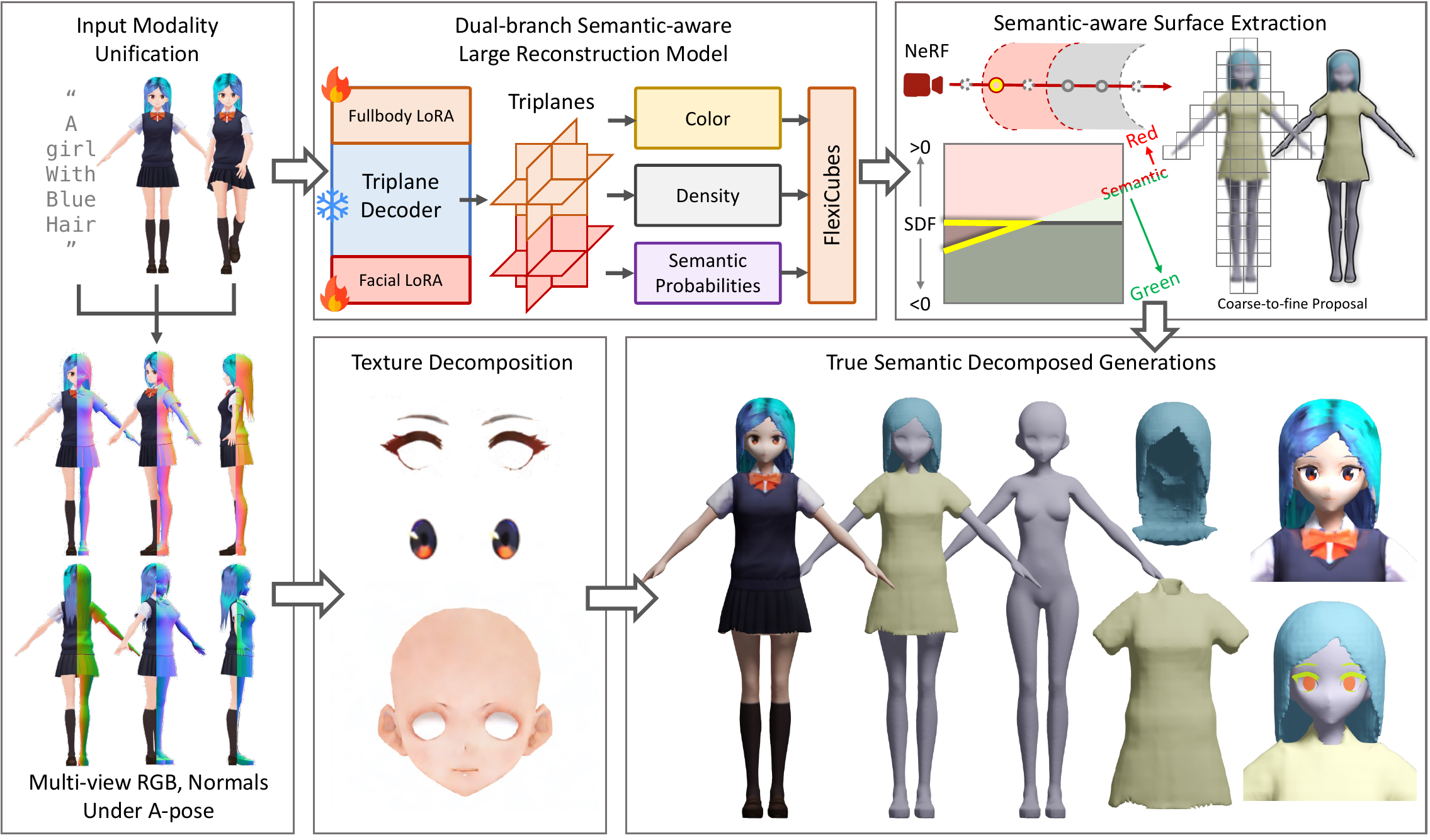}
\caption{Overview of the \textbf{StdGEN++} pipeline.
(1) \textbf{Input Modality Unification}: Diverse inputs (text or images) are first canonicalized into unified multi-view RGB and normal maps under A-pose.
(2) \textbf{Dual-branch S-LRM}: These inputs feed into our reconstruction model, which leverages specialized Fullbody and Facial LoRA branches to predict high-fidelity geometry and semantic fields.
(3) \textbf{Surface Extraction}: A semantic-aware extraction mechanism, accelerated by a coarse-to-fine proposal scheme, efficiently reconstructs high-resolution meshes from the implicit representations.
(4) \textbf{Texture Decomposition}: Finally, the system performs texture decomposition to separate appearance components.
Ultimately, the system yields structurally independent meshes (body, hollow clothing, hair) and editable texture layers.}
\label{fig:pipeline}
\end{figure*}

Crucially, Anime3D-EX enriches the original data with three specialized supervision signals to facilitate our dual-branch and decomposition training:
\begin{itemize}
\item \textbf{Hierarchical Semantic Renderings.} To support the S-LRM's layered reconstruction, we define three core semantic categories: (1) base minimal-clothed body, (2) clothing, and (3) hair. For each character, we generate multi-view renderings under three configurations: complete model, body with clothing, and base body alone.
\item \textbf{Head-Centric Facial Data.} To supervise the dedicated facial LoRA branch, we spatially crop and re-normalize the head region of each character. These head-centric assets are rendered with the same multi-layer semantic configurations as the full body, ensuring high-fidelity supervision for fine-grained facial geometry.
\item \textbf{Disentangled Texture \& Text.} For texture decomposition, we generate pixel-aligned ground-truth layers for the face, strictly isolating the \textit{eyebrow/lash}, \textit{base skin}, and \textit{iris}. Additionally, we utilize Qwen3-VL~\cite{Qwen3-VL} to generate rich, context-aware captions that describe the appearance and style of each character, enabling text-driven generation.
\end{itemize}

\section{Method}
\label{sec:method}

We present StdGEN++, a comprehensive system designed for the high-fidelity generation of semantically decomposed 3D characters.
The pipeline begins by unifying diverse input modalities into a canonical multi-view representation (Sec.~\ref{subsec:multiview}).
Taking these aligned multi-view images as input, we introduce the \textbf{Dual-branch S-LRM}, which reconstructs semantic-aware 3D geometry with specialized attention to facial fidelity (Sec.~\ref{subsec:lrm}).
To transform these predicted implicit representations into usable assets, we derive a novel formalism that explicitly extracts 3D surfaces corresponding to specific semantics, which is efficiently implemented via a coarse-to-fine proposal scheme to enable high-resolution output (Sec.~\ref{subsec:surface}).
The entire reconstruction network is supervised via a three-stage strategy incorporating photometric, geometric, and dedicated hole-filling regularization to ensure structural completeness (Sec.~\ref{subsec:train}). Complementing the geometric reconstruction, the pipeline includes a texture decomposition module (Sec.~\ref{subsec:tex}) that operates on the canonical view to separate appearance into editable layers. Finally, a selective multi-layer refinement process to polish surface quality (Sec.~\ref{subsec:refine}). An overview of the StdGEN++ pipeline is shown in Fig.~\ref{fig:pipeline}.

\subsection{Input Unification and Multi-view Generation}
\label{subsec:multiview}

Our pipeline establishes the \textbf{canonical A-pose character} as the standardized intermediate representation. This design choice minimizes self-occlusion and provides a consistent geometric basis for the subsequent S-LRM, decoupling the reconstruction complexity from input variations.
However, in practical character creation workflows, users rarely start with such standardized assets. Initial inputs are typically diverse and unconstrained, ranging from arbitrary-pose character illustrations to high-level textual descriptions.

To bridge the gap between diverse creative intents and standardized 3D reconstruction, we upgrade the canonicalization module into a \textbf{unified input framework}.
This framework supports three input modalities by mapping them onto the common A-pose interface:
(1) Direct A-pose images for standard assets;
(2) Arbitrary-pose images, which are re-targeted to A-pose while preserving identity;
(3) Pure text prompts, which are generated into visual canonical priors from scratch.
For cases (2) and (3), we integrate a specialized diffusion module built upon Stable Diffusion~\cite{rombach2022high} augmented with ReferenceNet~\cite{peng2024charactergen}.
\textbf{Unlike StdGEN, which focused primarily on image pose correction, this unified framework allows StdGEN++ to flexibly accept both visual and textual guidance}. This significantly broadens the system's applicability, ensuring that downstream geometry generation and texture decomposition can proceed uniformly regardless of the source modality.

\xhdr{A-pose Character Synthesis}
Given a text prompt or an arbitrary-pose reference (with or without text), our system synthesizes a canonical A-pose character image. When the input is purely textual, we use a fine-tuned Stable Diffusion model that directly generates A-pose character images from the text description, leveraging learned priors of human anatomy and artistic style. When an arbitrary-posed character image is provided, we employ a ReferenceNet-augmented diffusion model~\cite{peng2024charactergen} to re-render it in A-pose while preserving identity. In both cases, the output is a standardized A-pose image that serves as the unified entry point for subsequent multi-view generation.

\xhdr{Multi-view RGBs and Normals Generation}
From the synthesized (or directly provided) A-pose image, we generate six orthographic views (elevation $0^\circ$, azimuth ${-90^\circ, -45^\circ, 0^\circ, 45^\circ, 90^\circ, 180^\circ}$) of RGB and normal maps using an adapted Era3D~\cite{li2024era3d} framework. Leveraging memory-efficient row-wise attention across views and between RGB and normal branches, our implementation enforces geometric consistency and supports high-resolution output up to 1024×1024 through progressive training. Normals are generated jointly with RGBs, ensuring surface coherence across views.
Compared with CharacterGen~\cite{peng2024charactergen}, our choice can simultaneously generate high-resolution, multi-view consistent normal maps for mesh refinement. 
Besides, the two-step design allows for improved editing in the 2D A-pose space, facilitating the generation of decomposed characters for enhanced 3D editing applications.

\begin{figure}[tb]
\centering
\includegraphics[width=0.95\linewidth]{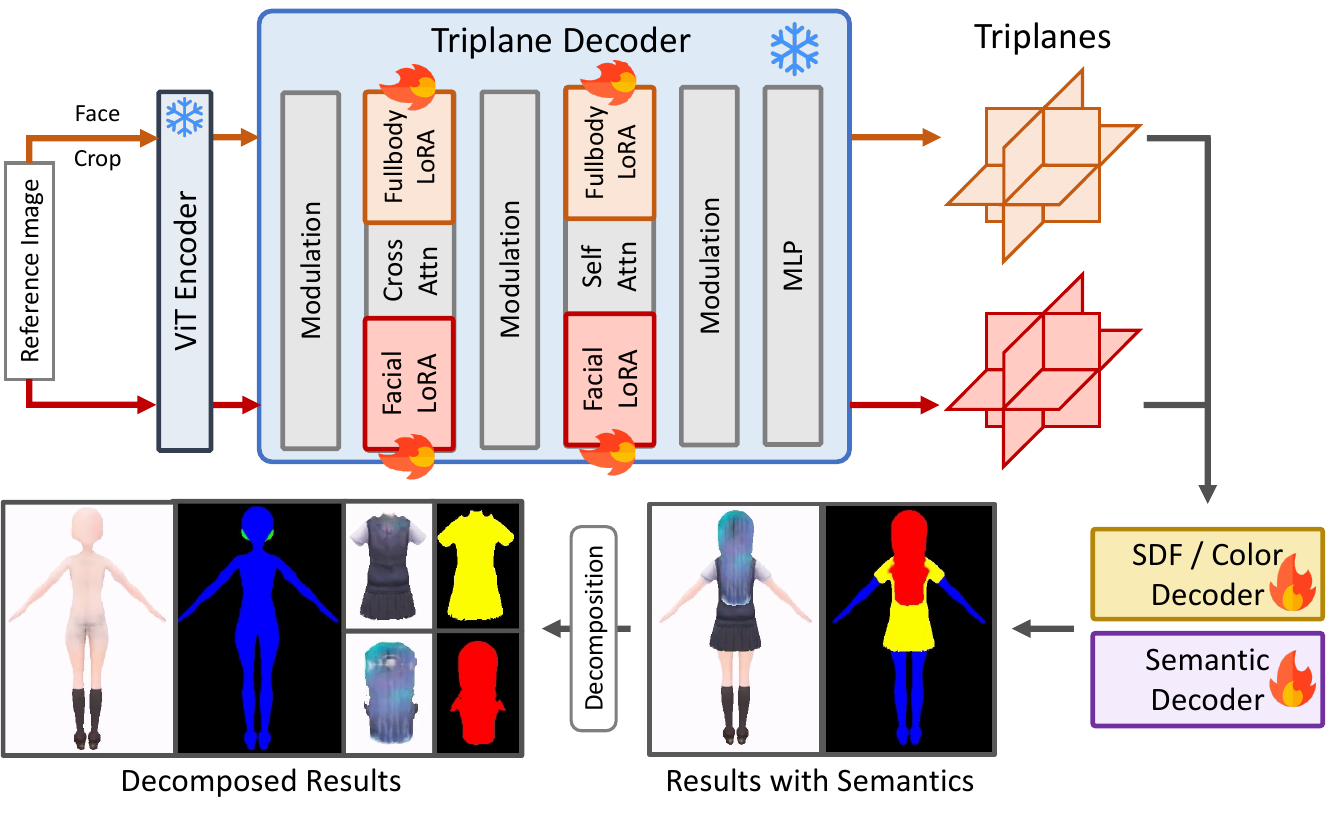}
\caption{Demonstration of the structure and intermediate outputs of our dual-branch semantic-aware large reconstruction model (S-LRM). }
\label{fig:lrm}
\end{figure}

\subsection{Dual-Branch Semantic-aware Large Reconstruction Model}
\label{subsec:lrm}
Once obtaining multi-view images,~\cite{xu2024instantmesh, peng2024charactergen} use transformer-based sparse-view Large Reconstruction Model~(LRM) to reconstruct a holistic 3D mesh without explicit semantic decomposition.

The success of StdGEN~\cite{he2025stdgen} demonstrates that extending the LRM framework with semantic awareness enables feed-forward reconstruction of decomposed 3D characters, separating body, clothing, and hair to support downstream applications in animation and game pipelines. Its core architecture follows InstantMesh~\cite{xu2024instantmesh}, consisting of a ViT encoder, an image-to-triplane transformer, and dedicated decoders for density/color and semantics.
However, StdGEN relies on a single reconstruction branch to handle the entire character. This monolithic processing creates an inherent bottleneck: owing to limited grid resolution and attention capacity, fine-grained details—particularly in the facial region—are often sacrificed to maintain global structure.

To overcome this limitation, we upgrade the architecture to a \textbf{dual-branch Semantic-aware Large Reconstruction Model (Dual-Branch S-LRM)} that significantly enhances reconstruction fidelity, particularly in facial regions critical for character believability. Unlike the single-branch design in StdGEN, our dual-branch system (Fig.~\ref{fig:lrm}) employs two specialized LoRA adapters~\cite{hu2021lora, qi2024tailor3d}: one processes full-body multi-view inputs to recover global structure and coarse semantics, while the other operates on cropped and resized head regions to capture fine-grained facial geometry and texture. Following prior practice~\cite{qi2024tailor3d}, we integrate LoRA modules into all linear layers within the self-attention and cross-attention blocks of the transformer, with each branch using its own set of trainable LoRA parameters.

Both branches follow the triplane NeRF/SDF paradigm: multi-view images are tokenized and fed into a transformer-based image-to-triplane decoder, whose output is decoded into semantic, color, and density/SDF fields. As in StdGEN, we adopt a two-stage training strategy—first optimizing via volume rendering on the NeRF representation, then refining with explicit mesh extraction using FlexiCubes~\cite{shen2023flexible} and rasterization-based losses.

By decoupling global and facial reconstruction into dedicated pathways, our dual-branch S-LRM achieves significantly higher fidelity in facial details while maintaining consistent overall structure. This addresses a key limitation of single-branch designs such as the original StdGEN, and better supports practical character creation scenarios.

\subsection{Semantic-aware Surface Extraction}
\label{subsec:surface}

To obtain a semantic-decomposed surface reconstruction, both NeRF and SDF implicit representations must be capable of rendering distinct semantic layers into images or extracting separate semantic surfaces using FlexiCubes in a differentiable manner.
To achieve that, a novel semantic-equivalent NeRF/SDF is proposed to extract character parts by specific semantics.

NeRF represents a 3D scene by spatial-variant volume densities with colors\footnote{We ignore the view-dependent effects to simplify the discussion.}.
We extend it with a semantic field, and model them as a learnable function $F_\Theta$ that takes sampled point location $\mathbf{x} = (x; y; z)$ as inputs, and outputs color $c$, density $\sigma$ and semantic distribution $s$ as:
$(\sigma, c, s) = F_\Theta(\mathbf{x})$.

To render per-pixel color $\hat{C}(\mathbf{r})$, a series of 3D points are sampled along the ray $\mathbf{r}$, and the pixel color is computed by integrating the sampled densities $\sigma_i$ and colors $c_i$ using the volume rendering equation with:
\begin{align} 
\hat{C}(\boldsymbol{r}) &=\sum_{i=1}^N T_{i}\alpha_i c_i,\ T_{i}=\prod_{j=1}^{i-1}(1-\alpha_j),
\end{align}
where $\alpha_i=(1-\exp (-\sigma_i \delta_i))$, $\delta_i=t_{i+1}-t_i$ is the alpha value of samples and the distance between adjacent samples. 

Given the probability $p_{s,i}$ of semantic $s$ at location $i$, the pixel color $\hat{C}_s(\mathbf{r})$ under semantic $s$ can be calculated as:
\begin{align} 
\hat{C}_s(\boldsymbol{r}) &=\sum_{i=1}^N T_{s,i}p_{s,i}\alpha_i c_i,\ T_{s,i}=\prod_{j=1}^{i-1}(1-\alpha_jp_{s,j}),
\end{align} 

If the probability of a certain semantic at a given location is zero, it should be considered fully transparent under the current semantic category. Furthermore, given that a position is known to be opaque, the probability of the current semantics should be linear to the final equivalent transparency.

\begin{figure}[tb]
\includegraphics[width=1.0\linewidth]{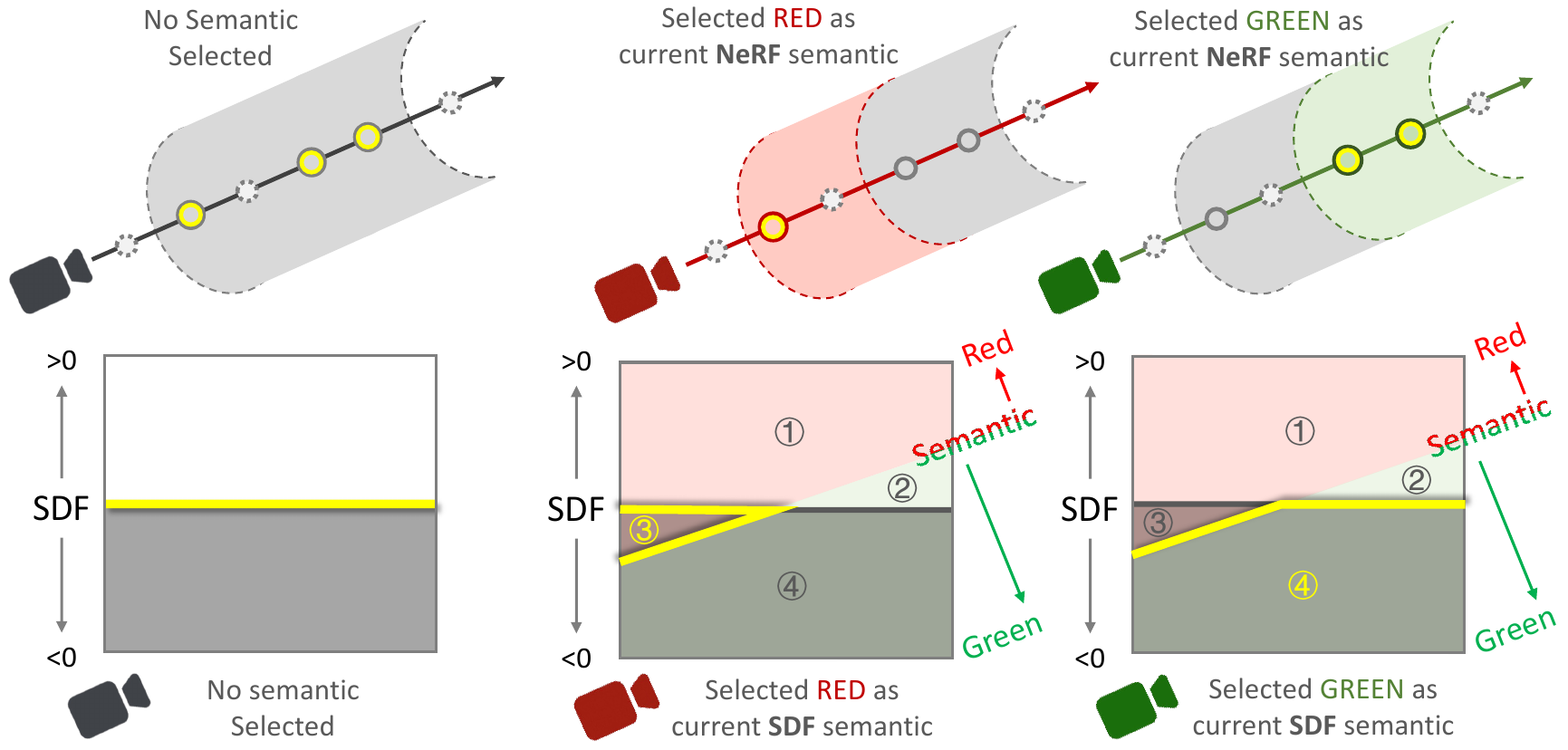}
\caption{Our semantic-equivalent NeRF and SDF extraction scheme (shown in yellow color).}
\label{fig:semantic}
\end{figure}

Unlike NeRF, SDF does not incorporate the concept of transparency.
Instead, positive/negative values represent points outside/inside the surface. 
Consequently, semantic probabilities cannot be directly applied to SDF for the mesh part extraction. 
Upon analysis, the extraction of a semantic-equivalent SDF should adhere to the following principles:

\begin{enumerate}
    \item The zero value of the original SDF serves as a hard constraint. When the original SDF is positive, the equivalent SDF should also be positive;
    \item When the original SDF is negative, the equivalent SDF should be zero at the boundaries where the maximum of relevant semantics transits;
    \item At locations where the original SDF equals zero, but the probability of the current semantic is not the highest among all semantics, the equivalent SDF should not only maintain its sign but also be greater than zero.
\end{enumerate}

Based on these principles, we propose the following formula for constructing the equivalent SDF:
\begin{align}
    f_{i,s}=\max(f_i, (\max_{r\neq s}p_{i,r})-p_{i,s} ),
\end{align}
Where $f_{i}$, $f_{i,s}$ are the original SDF and equivalent SDF of semantic $s$ at location $i$, respectively.
Fig.~\ref{fig:semantic} illustrates our method's scheme. For red semantics, only region 3 is selected, as regions 1, 2 (SDF$>$0) and region 4 (non-red) are discarded. Similarly, when green is chosen, region 4 is correctly extracted.
This formulation ensures correct decomposition by specific semantics and is fully compatible with subsequent FlexiCubes mesh extraction. In this way, we can differentially extract multi-layer semantic surfaces from S-LRM's outputs, greatly facilitating the LRM training and downstream optimization.

\textbf{Surpassing the resolution constraints of StdGEN} ($100 \times 100 \times 150$)~\cite{he2025stdgen}, we aim to extract high-fidelity geometric details at a significantly scaled-up resolution of $256 \times 256 \times 384$. However, directly applying the original dense evaluation strategy at this scale would incur prohibitive memory and computational costs. To address this, we introduce a \textbf{novel coarse-to-fine proposal scheme} that restricts the heavy network evaluations to a sparse set of active voxels.

Let $\mathcal{V}_{L}$ and $\mathcal{V}_{H}$ denote the vertex sets of the low-resolution coarse grid and the target high-resolution grid, respectively. We first compute the coarse SDF values $f^c$ on $\mathcal{V}_{L}$. The region of interest is determined by identifying the implicit surface boundary, enhanced by a morphological dilation to ensure coverage. Formally, we define the binary occupancy mask $M_{L}$ on the coarse grid as:
\begin{align}
    M_{L}(\mathbf{v}) = \max_{\mathbf{u} \in \mathcal{N}_k(\mathbf{v})} \mathbb{1}\left(f^c(\mathbf{u}) < 0\right), \quad \forall \mathbf{v} \in \mathcal{V}_{L},
\end{align}
where $\mathbb{1}(\cdot)$ is the indicator function, and $\mathcal{N}_k(\mathbf{v})$ represents the $k \times k \times k$ spatial neighborhood (kernel size $k=3$) centered at $\mathbf{v}$. This operation effectively dilates the surface boundary, providing a safety margin for subsequent operations.

The mask is then upsampled to the high-resolution space via nearest-neighbor interpolation $\mathcal{U}(\cdot)$, defining the active computational domain $\Omega_{active}$:
\begin{align}
    \Omega_{active} = \{ \mathbf{p} \in \mathcal{V}_{H} \mid \mathcal{U}(M_{L})(\mathbf{p}) = 1 \}.
\end{align}
Finally, the fine-grained predictions for SDF, deformation, and semantics are exclusively executed on vertices within $\Omega_{active}$. This reduces complexity from $\mathcal{O}(|\mathcal{V}_{H}|)$ to $\mathcal{O}(|\Omega_{active}|)$, where $|\Omega_{active}| \ll |\mathcal{V}_{H}|$, thereby enabling high-resolution reconstruction with manageable resource consumption.

\subsection{Semantic-aware Training Scheme}
\label{subsec:train}

Current LRMs typically rely solely on 2D supervision, which limits their ability to generate information about objects' internal structures under occlusion; 3D supervision would be effective but often too resource-intensive. 
To address this, we propose an effective supervision that jointly learns semantics and colors, enabling the acquisition of a 3D semantic field and internal character information using only 2D supervision. 

\xhdr{Stage 1: Training on NeRF with Single-layer Semantics}
In this initial stage, we train on the triplane NeRF representation.
We initialize the model with the pre-trained InstantNeRF, training the newly added LoRA in all attention blocks' linear layers and the newly introduced semantic decoder.
We train it under the image, mask, and semantic loss:
\begin{align}
    & \hat{\mathcal{S}}(\boldsymbol{r}) =\sum_{i=1}^N T_ip_i\alpha_i,\
    \mathcal{L}_{\text{sem}} = \sum_k CE(\hat{\mathcal{S}_k}, \mathcal{S}^{gt}_k),
    \\ & \mathcal{L}_1 = \mathcal{L}_{\text{mse}} + \lambda_{\text{lpips}} \mathcal{L}_{\text{lpips}} + \lambda_{\text{mask}}\mathcal{L}_{\text{mask}} + \lambda_{\text{sem}}\mathcal{L}_{\text{sem}},
\end{align}
$\hat{\mathcal{S}}$ is the semantic map calculated by the probabilities $p_{i}$ from semantic decoder's output through a softmax layer.
$\hat{\mathcal{S}_k}, \mathcal{S}^{gt}_k$ denotes the $k$-th view of rendered and ground-truth semantic maps, and $CE$ denotes the cross-entropy function.

\xhdr{Stage 2: Training on NeRF with Multi-layer Semantics}
Having learned robust surface semantic information in the first stage, we aim to learn the 3D character's internal semantic and color information.
We hierarchically supervise from outside to inside according to the spatial relationship of different semantic parts, by masking specific semantics during rendering and supervising with corresponding 2D ground truth.
Assuming we aim to preserve a set of semantics $\{P_s\}$, we can render the image and semantic map under current conditions as follows:
\begin{align}
    \hat{C}_P(\boldsymbol{r}) &=\sum_{i=1}^N T_{P,i}\alpha_i c_i\sum_{s\in P}p_{s,i}, \\
    \hat{\mathcal{S}}_P(\boldsymbol{r}) &=\sum_{i=1}^N T_{P,i}\alpha_i p_i\sum_{s\in P}p_{s,i}, \\
    \mbox{where }\ T_{P,i} &=\prod_{j=1}^{i-1}(1-\alpha_j\sum_{s\in P}p_{s,j}),
\end{align}
The loss function is defined as:
\begin{align}
\mathcal{L}_2 & = \mathcal{L}_{\text{mse},P} + \lambda_{\text{lpips}} \mathcal{L}_{\text{lpips},P} + \lambda_{\text{mask}}\mathcal{L}_{\text{mask},P}\nonumber \\
&+ \lambda_{\text{sem}}\sum_k CE(\hat{\mathcal{S}}_{P,k}, \mathcal{S}^{gt}_{P,k}),
\end{align}
This decomposed training approach enables our S-LRM to simultaneously learn color and semantic information for the surface and the object's interior, thus achieving feed-forward 3D content decomposition and reconstruction.

\xhdr{Stage 3: Training on Mesh with Multi-layer Semantics}
We switch to mesh representation~\cite{shen2023flexible}
for efficient high-resolution training.
We then extract the equivalent SDF via:
\begin{align}
    f_{i,P}=\max(f_i, (\max_{s\notin P}p_{i,s}-\max_{s\in P}p_{i,s}) ),
\label{eq:semantic_sdf}
\end{align}
Subsequently, we input the equivalent SDF into FlexiCubes to obtain the mesh, render the image and semantic map, and supervise using the following loss function:
\begin{align}
    \mathcal{L}_3 = \mathcal{L}_2 &+ \lambda_{\text{normal}} \sum_k M^{gt}_P\otimes\left(1 - \hat{N}_{P,k}\cdot N_{P,k}^{gt}\right) \nonumber\\
 &+ \lambda_{\text{depth}} \sum_{k} M^{gt}_P\otimes\left\|\hat{D}_{P,k}-D_{P,k}^{gt}\right\|_1 \nonumber\\
 &+ \lambda_{\text{dev}}\mathcal{L}_{\text{dev}} + \lambda_{\text{hole}}\mathcal{L}_{\text{hole},P'},
\end{align}
where $\hat{D}_{P,k}$, $\hat{N}_{P,k}$, denotes the rendered depth and normal; $D_{P,k}^{gt}$, $ N_{P,k}^{gt}$ and $M_P^{gt}$ denote the ground truth depth, normal, and mask of the $k$-th view under semantic set $P$, respectively; $\mathcal{L}_{\text{dev}}$ denotes the deviation loss of FlexiCubes. 

\textbf{To address the topological fracturing often observed in thin structures (e.g., clothing) in StdGEN~\cite{he2025stdgen}, we introduce a dedicated hole-filling regularization, denoted as $\mathcal{L}_{\text{hole},P'}$.}
This term is specifically applied to semantic subsets $P'$ prone to topological holes due to the sign-sensitive nature of SDF-based extraction.
Let $f_{i,P'}$ denote the semantic-aware equivalent SDF for region $P'$, as defined in Eq.~\eqref{eq:semantic_sdf}. We define $\vec{\mathcal{E}}_{P'}$ as the set of all directed edges $(f_a, f_b)$ between adjacent grid vertices $(a, b)$ such that $f_a > 0$ and $f_b < 0$. The hole-filling loss is then given by applying the sign-stabilization objective to these edges:
\begin{align}
\mathcal{L}_{\text{hole},P'} := \sum_{(f_a, f_b) \in \vec{\mathcal{E}}_{P'}} H\big(\sigma(f_a),\, \mathrm{sign}(f_b)\big),
\end{align}
where $\sigma(\cdot)$ is the sigmoid function, $\mathrm{sign}(\cdot)$ returns $\pm1$, and $H(p, q) = -[q \log p + (1-q) \log(1-p)]$ is the binary cross-entropy loss.
This formulation gently pulls positive SDF values within the semantic-aware representation for \(P'\) toward neighboring negative regions across thin structures, thereby preserving interior (\(f < 0\)) connectivity while retaining the sign change necessary for surface definition.

\subsection{Texture Decomposition}
\label{subsec:tex}
In practical character production pipelines, particularly in animation, gaming, and virtual avatars, textures should support part-wise editing, expression control, and gaze tracking. \textbf{A fundamental limitation of StdGEN~\cite{he2025stdgen} and most existing approaches is the generation of a monolithic texture atlas}. This representation entangles semantically distinct components such as skin, hair, eyebrows, and eyes into a single image. This coupling prevents independent manipulation (e.g., changing iris color without affecting sclera) and complicates integration with rigging or eye-tracking systems.

To overcome this fundamental limitation, \textbf{we introduce a novel semantic texture decomposition paradigm}. Our approach assigns each anatomical component to its own dedicated texture map. Distinct from simple segmentation, our key insight is to formulate this decomposition as a generative multi-frame inpainting problem. Inspired by video diffusion frameworks~\cite{hu2024animate}, we train a model where each output frame corresponds to a predefined semantic region, such as the eyebrow, iris, or base skin. The input is the original unified texture rendered from a canonical front-facing view, which serves as a geometrically aligned reference for decomposition—sufficient for facial regions due to their near-frontal visibility and symmetry in standard character designs.
Internally, our model employs spatial and temporal attention mechanisms across both feature layers and frames. This enables information exchange not only within each part (via spatial attention) but also between different semantic regions (via temporal attention), ensuring visual consistency while allowing structural separation. 

We instantiate this framework on facial textures, following industry-standard layering practices observed in high-fidelity anime assets. The output is structured as a three-frame video:
\begin{itemize}
\item Frame 1: combined \textit{eyebrow and eyelash} layer (non-overlapping, further separable via connectivity masks);
\item Frame 2: \textit{base skin} with face and eye white;
\item Frame 3: \textit{eye iris} with pupil and specular highlights.
\end{itemize}
This hierarchy mirrors real-world production workflows, where iris and skin are always separated to enable gaze redirection and dynamic wetness effects.

During training, we simulate application-specific perturbations to improve robustness and facilitate integration with our framework. Each training sample undergoes one or both of the following augmentations independently with a 50\% probability each: (1) an oil-painting stylization to mimic artistic variation; or (2) re-rendering of the source 3D model under a random pose, followed by A-pose canonicalization by diffusion model in Sec.~\ref{subsec:multiview}, and cropping to the canonical facial region, ensuring the decomposition remains stable in actual pipelines.
The resulting decomposed textures are not only visually faithful but also directly editable—enabling applications such as iris recoloring, brow reshaping, or eye tracking without reprocessing the full character.

\subsection{Multi-layer Refinement}
\label{subsec:refine}

While our upgraded S-LRM directly yields high-fidelity geometry with sharp details, distinct semantic parts may benefit from tailored post-processing strategies. Recent methods~\cite{wu2024unique3d, li2024craftsman} utilizing high-resolution normal maps for mesh optimization have shown promising results, albeit primarily for holistic meshes. We propose an iterative optimization mechanism for multi-layer mesh refinement.

To prevent inter-penetration during optimization, we employ a staged approach: Initially, we optimize the base minimal-clothed human model; subsequently, outer layers (clothing and hair) can be sequentially optimized while treating the inner layers as fixed collision boundaries.
The optimization process is guided by the multi-view normal maps generated via the diffusion module. Each step involves differentiable rendering to compute gradients for vertex adjustments and re-meshing operations (edge collapse, split, and flips). The loss function is defined as:
\begin{align}
    \mathcal{L}_{r1}=\ & \lambda_{\text{mask}}' \sum_k || \hat{M}_k-M_k^{\text{pred}} ||_2^2 +\ \lambda_{\text{col}} \mathcal{L}_{\text{col}} \nonumber \\
    +\ & \lambda_{\text{normal}}' M_k^{\text{pred}} \otimes \sum_k || \hat{N}_k-N_k^{\text{pred}} ||_2^2
\end{align}
where $\hat{M}_k$, $\hat{N}_k$ are rendered masks and normal maps, $M_k^{\text{pred}}$, $N_k^{\text{pred}}$ are diffusion-generated masks and normal maps under $k$-th view, respectively. $\mathcal{L}_{\text{col}}$ is the collision loss modified from \cite{peng2024pica} to ensure outer-layer mesh does not penetrate the inner-layer mesh:
\begin{align}
    \mathcal{L}_{\text{col}}=\frac{1}{n}\sum_{i=1}^n\max{((v_j-v_i)\cdot n_j, 0)^3}
\end{align}
where $v_i$ represents the $i$-th vertex of the outer-layer mesh, $v_j$ is its nearest neighbor of $v_i$ on the inner-layer mesh, and $n_j$ denotes the normal vector associated with $v_j$.
Upon completing the optimization process, the mesh undergoes an additional ExplicitTarget Optimization phase, similar to that employed in Unique3D~\cite{wu2024unique3d}. This stage aims to eliminate multi-view inconsistencies and further refine the geometry. Finally, the optimized meshes are colorized by the back projection of the multi-view images.

It is worth noting that thanks to the high-resolution capability of our proposed S-LRM, this refinement step is primarily deployed for the body layer to enhance skin smoothness, while it can be optionally bypassed for cloth and hair to preserve their sharp, thin structures generated by the primary reconstruction network.

\section{Experiments}
\label{sec:experiments}

\subsection{Implementation Details}
We adopt the dataset settings from StdGEN, partitioning the data into training and testing sets with a 99:1 ratio. For the diffusion components, we employ a progressive resolution training strategy. The canonicalization diffusion model is initially trained at a $512$ resolution with a learning rate of $5\times 10^{-5}$, which is subsequently reduced to $1\times 10^{-5}$ as the resolution scales up to $768$ and $1024$. Conversely, the multi-view diffusion model maintains a constant learning rate of $5\times 10^{-5}$ while progressively scaling across resolutions of $512$, $768$, and $1024$. The video diffusion model for texture decomposition operates at a resolution of $512\times 512$.

For the dual-branch S-LRM, we integrate Low-Rank Adaptation (LoRA)~\cite{hu2021lora} into the transformer architecture, specifically modifying the query, key, and value projection layers within both self-attention and cross-attention modules. We set the LoRA rank to $128$ for each branch and train with a learning rate of $4\times 10^{-5}$. Following InstantMesh~\cite{xu2024instantmesh}, the model takes $6$ multi-view RGB images at a resolution of $320 \times 320$ as input. During inference, inputs for the facial branch are specifically obtained by cropping the face region from the generated multi-view images and resizing them to $320\times 320$. The training process encompasses three supervision stages with rendering resolutions of $192$, $144$, and $512$, respectively. The loss weights are configured as follows: $\lambda_{\text{lpips}}=2.0, \lambda_{\text{mask}}=1.0, \lambda_{\text{sem}}=1.0, \lambda_{\text{depth}}=0.5, \lambda_{\text{normal}}=0.2, \lambda_{\text{dev}}=0.5$, and $\lambda_{\text{hole}}=10^{-4}$. Notably, to enforce higher precision on facial features, $\lambda_{\text{mask}}$ is increased to $10.0$ for the facial branch.

For geometry extraction via FlexiCubes, we configure the sampling grid dimensions and physical scales distinctively for each branch to match their respective scopes. The full-body branch utilizes a grid size of $256\times 256\times 384$ spanning a volume of $0.7\times 0.7\times 1.05$ (relative to the bounding unit cube of the character). The facial-specific branch employs a grid size of $180\times 180\times 180$ within a volume of $0.25\times 0.25\times 0.25$.

\begin{figure*}[htbp]
\centering
\includegraphics[width=0.90\linewidth]{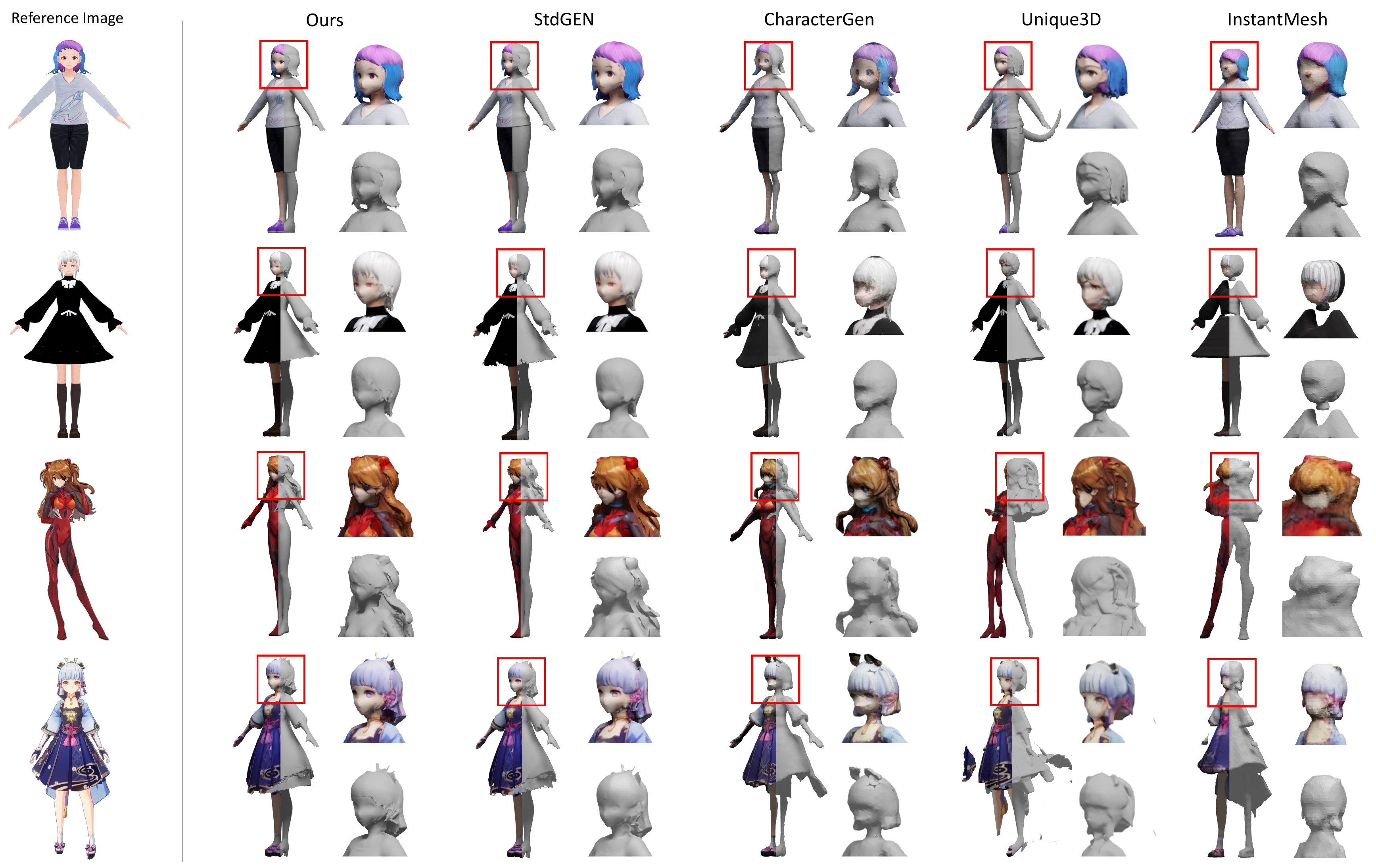}
\caption{Qualitative comparisons on geometry and appearance of generated 3D characters.}
\label{fig:qual}
\end{figure*}

\subsection{Holistic Generation Comparisons}
Since existing baselines lack the capability for layered 3D generation, we focus our comparative analysis on the holistic (non-layered) generation results. We conduct evaluations on the test split of the Anime3D++ dataset.
To ensure a fair comparison regarding pose variation, we decouple the pose canonicalization component and evaluate two distinct scenarios: (1) A-pose inputs, where all methods are compared against A-pose ground truth; and (2) Arbitrary pose inputs. For the latter, following the protocol established in CharacterGen~\cite{peng2024charactergen}, we compare our method and CharacterGen (both capable of canonicalization) against the A-pose ground truth, while other methods are compared against the ground truth in the original input pose.

\xhdr{Baselines and Metrics}
We benchmark against a diverse set of state-of-the-art approaches. For 2D multi-view generation, we compare with Zero-1-to-3~\cite{liu2023zero}, SyncDreamer~\cite{liu2023syncdreamer}, Era3D~\cite{li2024era3d}, and CharacterGen~\cite{peng2024charactergen}. For 3D character generation, baselines include SDS-based optimization methods (Magic123~\cite{qian2023magic123}, ImageDream~\cite{wang2023imagedream}), feed-forward methods (OpenLRM~\cite{hong2023lrm,openlrm}, LGM~\cite{tang2024lgm}, InstantMesh~\cite{xu2024instantmesh}), and direct mesh reconstruction methods (Unique3D~\cite{wu2024unique3d}).
We employ standard metrics including SSIM~\cite{wang2004image}, LPIPS~\cite{zhang2018perceptual}, and FID to evaluate generation quality and perceptual fidelity. Additionally, we compute the CLIP~\cite{radford2021learning} cosine similarity between the reference image and the generated views (or renderings) to assess semantic consistency. For 3D evaluations, results are rendered as eight equidistant azimuth views at zero elevation and aligned via horizontal mask registration.

\newcolumntype{C}[1]{>{\centering\arraybackslash}m{#1}}

\begin{table*}[htb]
    \centering
    \caption{Quantitative comparison of A-pose and arbitrary pose inputs for 2D multi-view generation and 3D character generation.}
    \resizebox{0.9\linewidth}{!}{
    \begin{tabular}{clcccc|cccc}
        \toprule
        & & \multicolumn{4}{c}{A-pose Conditioned Input} & \multicolumn{4}{c}{Arbitrary-pose Conditioned Input} \\
        \cmidrule(lr){3-6} \cmidrule(lr){7-10}
        & & SSIM$\uparrow$ & LPIPS$\downarrow$ & FID$\downarrow$ & CLIP Similarity$\uparrow$ & SSIM$\uparrow$ & LPIPS$\downarrow$ & FID$\downarrow$ & CLIP Similarity$\uparrow$ \\
        \midrule
        \multirow{5}{*}{\textbf{\shortstack{Multi-view \\ Comparisons \\ in 2D}}} & SyncDreamer~\cite{liu2023syncdreamer}   & 0.870	& 0.183	& 0.223	& 0.864	& 0.845	& 0.217	& 0.328 &	0.839 \\
        & Zero-1-to-3~\cite{liu2023zero}       & 0.865	& 0.172	& 0.500	& 0.885	& 0.842	& 0.209	& 0.481 & 	0.878 \\
        & Era3D~\cite{li2024era3d}         & 0.876	& 0.144	& 0.095	& 0.908	& 0.842	& 0.195	& 0.094 & 	0.900 \\
        & CharacterGen~\cite{peng2024charactergen}  & 0.886	& 0.119	& 0.063	& 0.928	& 0.871	& 0.139	& 0.056 & 	0.919 \\
        & \textbf{Ours}         & \textbf{0.958}	& \textbf{0.038}	& \textbf{0.004}	& \textbf{0.941}	& \textbf{0.920}	& \textbf{0.071}	& \textbf{0.014} &	\textbf{0.935} \\
        \midrule
        \multirow{9}{*}{\textbf{\shortstack{Character \\ Comparisons \\ in 3D}}} & Magic123~\cite{qian2023magic123}      & 0.886	& 0.142	& 0.192	& 0.887	& 0.849	& 0.197	& 0.256 &	0.862 \\
        & ImageDream~\cite{wang2023imagedream}    & 0.856	& 0.171	& 0.846	& 0.836	& 0.823	& 0.218	& 0.875 &	0.818 \\
        & OpenLRM~\cite{openlrm}       & 0.889	& 0.151	& 0.406	& 0.878	& 0.863	& 0.191	& 0.707 &	0.844 \\
        & LGM~\cite{tang2024lgm}           & 0.876	& 0.151	& 0.282	& 0.902	& 0.838	& 0.203	& 0.480 &	0.884 \\
        & InstantMesh~\cite{xu2024instantmesh}   & 0.888	& 0.126	& 0.107	& 0.906	& 0.846	& 0.202	& 0.285 &	0.886 \\
        & Unique3D~\cite{wu2024unique3d}      & 0.889	& 0.136	& 0.030	& 0.919	& 0.856	& 0.190	& 0.042 &	0.903 \\
        & CharacterGen~\cite{peng2024charactergen}  & 0.880	& 0.124	& 0.081	& 0.905	& 0.869	& 0.134	& 0.119 &	0.901 \\
        & StdGEN          & 0.937	& 0.066	& \textbf{0.010}	& \textbf{0.941}	& \textbf{0.916}	& \textbf{0.084}	& \textbf{0.011} &	0.936 \\
        & \textbf{Ours (StdGEN++)} & \textbf{0.938} & \textbf{0.064} & 0.011 & \textbf{0.941} & \textbf{0.916} & \textbf{0.084} & \textbf{0.011} & \textbf{0.937} \\
        \bottomrule
    \end{tabular}
    }
    \label{tb:main}
\end{table*}

\xhdr{Quantitative Results}
As presented in Tab.~\ref{tb:main}, our method demonstrates consistent superiority across both standard and arbitrary pose settings.
Existing 2D multi-view methods often fail to maintain 3D geometric consistency, leading to inferior scores. Among 3D baselines, SDS-based approaches typically suffer from blurred geometry and the Janus problem, while feed-forward methods generally trade geometric precision for speed.
Notably, while Unique3D achieves competitive metrics due to high-resolution supervision, it suffers from unstable mesh initialization, which compromises robustness. CharacterGen shows advantages in arbitrary pose settings due to its canonicalization capability; however, its performance diminishes significantly in A-pose tasks, indicating limited reconstruction fidelity.
In contrast, our method outperforms all baselines, achieving the best balance between geometric accuracy, texture fidelity, and semantic consistency. Furthermore, comparisons with StdGEN reveal that StdGEN++ maintains consistent performance, with slight improvements in perceptual metrics (e.g., A-pose LPIPS reduced from 0.066 to 0.064).

\xhdr{Qualitative Results}
Visual comparisons in Fig.~\ref{fig:qual} highlight the distinct advantages of our approach.
Current SOTA methods exhibit several limitations: InstantMesh is heavily constrained by its grid resolution, resulting in over-smoothed textures and missing details. Unique3D, despite its high resolution, relies heavily on depth estimation; inaccuracies in predicted depth frequently lead to severe geometric collapse or distortions. CharacterGen, while handling poses well, often produces low-fidelity textures and is plagued by visually disruptive black artifacts during back-projection.
Conversely, our method produces sharp, artifact-free geometries with superior texture details. Even under complex pose inputs, our model successfully recovers the canonical shape with high fidelity, significantly surpassing competing methods in visual quality.

\subsection{Decomposed Geometry Evaluation Between StdGEN++ and StdGEN}
\begin{figure}[htbp]
\centering
\includegraphics[width=1\linewidth]{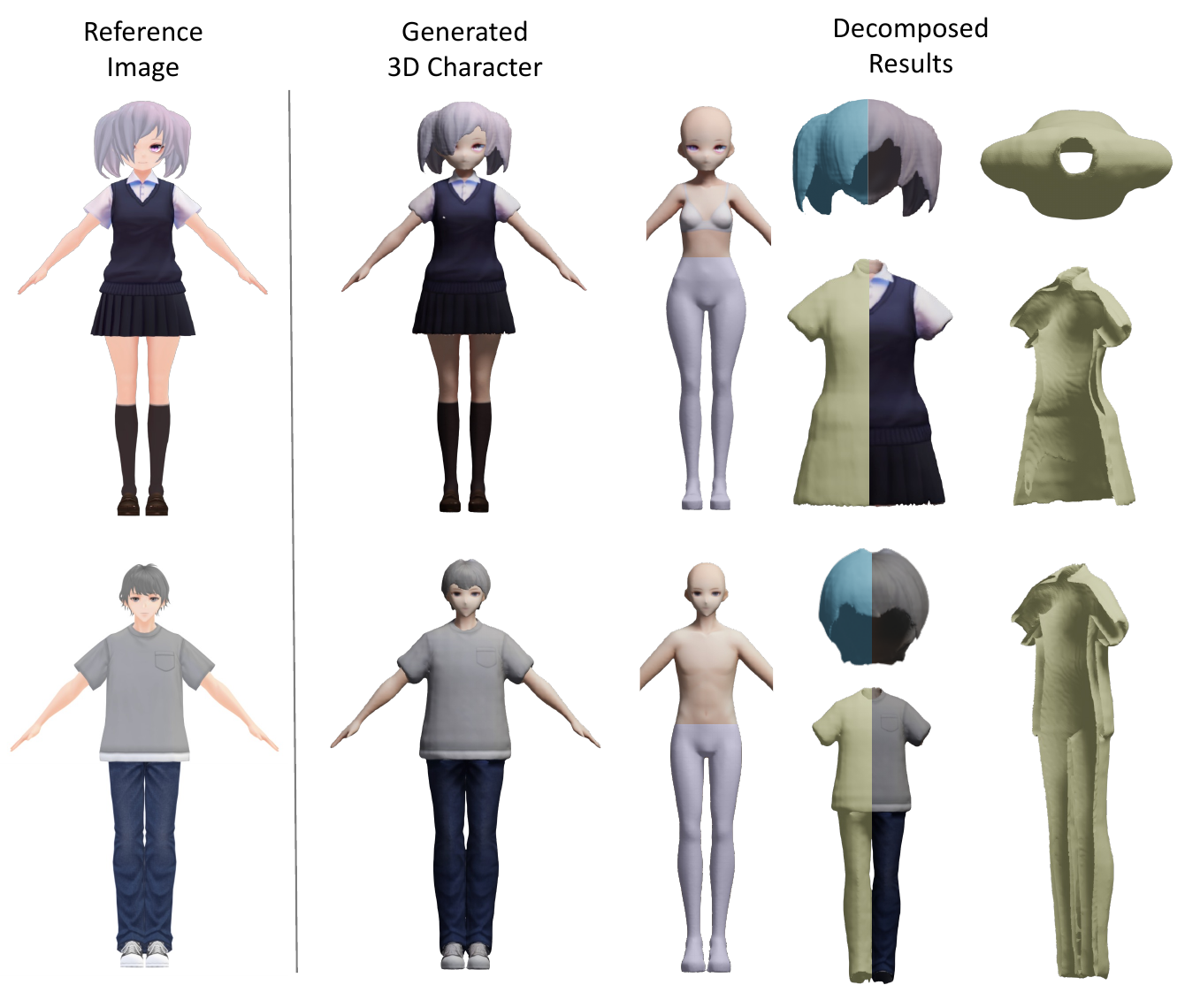}
\caption{\textbf{Layered decomposition results.}
From left to right: input reference, generated holistic character, and semantically decomposed layers (body, cloth, hair). The cross-sectional views (rightmost column) reveal that the reconstructed clothing is accurately modeled with internal hollow structures, ready for physics simulation.}
\label{fig:decomp_vis}
\end{figure}

Unlike holistic generation, our framework is designed as a comprehensive system that uniquely supports high-fidelity layered decomposition. As illustrated in Fig.~\ref{fig:decomp_vis}, the system successfully decouples the character into independent semantic layers (body, clothing, and hair) while maintaining high geometric fidelity. Uniquely, our approach generates the clothing as a standalone, internally hollow mesh (see the cross-sectional views in Fig.~\ref{fig:decomp_vis}). This structural independence is critical for industrial pipelines, enabling downstream applications like realistic cloth simulation and collision handling that are unattainable with non-layered surface generation.

To quantitatively evaluate this capability, we compare the geometric accuracy of each decomposed layer (Body, Cloth, Hair) against the ground truth meshes\footnote{For layered evaluation, we exclude 8 samples from the original 109 test cases due to ambiguous or defective ground-truth semantic labels, which would render layer-wise metrics mathematically invalid.}.

\xhdr{Layered Reconstruction Quality}
To provide a comprehensive assessment of geometric fidelity, we employ three complementary metrics: Chamfer Distance (CD) for surface accuracy (lower is better); Volumetric IoU (evaluated at $1/32$ granularity) for volumetric consistency (higher is better); and F-Score (F1$^{0.5}$) with a strict threshold of $\tau = 0.5\%$ to assess fine-scale alignment (higher is better).

Tab.~\ref{tb:decompose} reports the quantitative comparison between StdGEN and the proposed StdGEN++.
By integrating the coarse-to-fine proposal scheme into the robust system architecture, StdGEN++ achieves consistent improvements across all semantic layers.
Notably, for the \textit{Hair} layer—the most geometrically complex component—our system improves the F1$^{0.5}$ score drastically from $0.642$ to $0.725$. This indicates that the upgraded pipeline successfully captures fine hair structures under the strict $0.5\%$ error threshold.

\begin{figure}[htbp]
\centering
\includegraphics[width=1\linewidth]{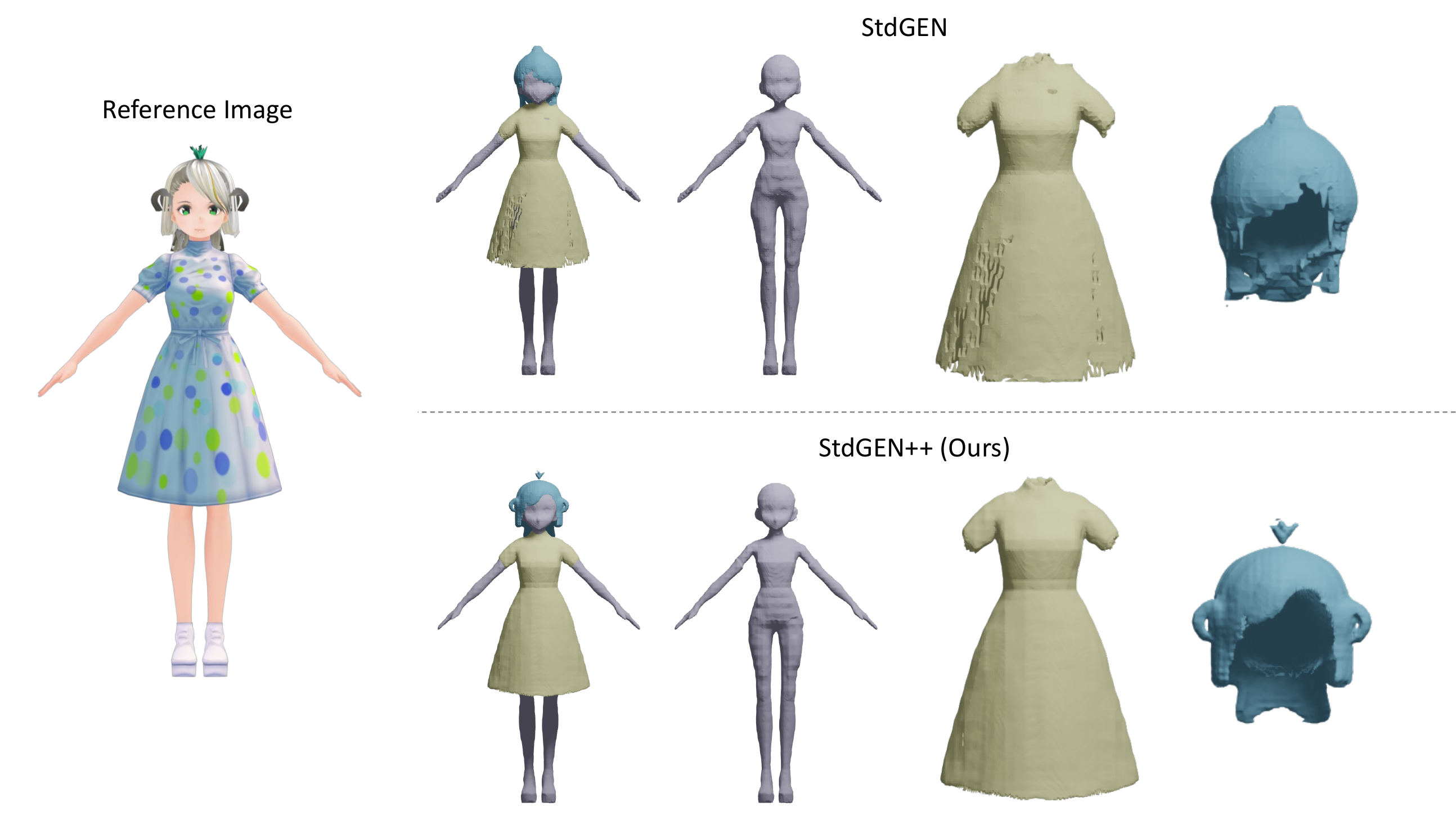}
\caption{\textbf{Visual comparison with the StdGEN.}
The baseline often suffers from topological artifacts due to low resolution, such as fractured skirts and loss of high-frequency details (e.g., hair strands). In contrast, our method (StdGEN++) directly produces coherent meshes with fine geometric details without any post-processing.}
\label{fig:comp_stdgen}
\end{figure}

\begin{table}[t]
    \centering
    \caption{Quantitative comparison of decomposed geometry quality. Ours (StdGEN++) significantly outperforms the StdGEN baseline. The substantial gain in F1$^{0.5}$ highlights our superior precision in recovering fine details.}
    \resizebox{\linewidth}{!}{
    \begin{tabular}{l|ccc|ccc}
        \toprule
        & \multicolumn{3}{c|}{StdGEN} & \multicolumn{3}{c}{\textbf{Ours (StdGEN++)}} \\
        \cmidrule(lr){2-4} \cmidrule(lr){5-7}
        Layer & CD$\downarrow$ & Voxel IoU$\uparrow$ & F1$^{0.5}$$\uparrow$ & CD$\downarrow$ & Voxel IoU$\uparrow$ & F1$^{0.5}$$\uparrow$ \\
        \midrule
        Body  & 0.0404 & 0.4738 & 0.654 & \textbf{0.0357} & \textbf{0.5058} & \textbf{0.690} \\
        Cloth & 0.0480 & 0.4345 & 0.605 & \textbf{0.0422} & \textbf{0.4682} & \textbf{0.644} \\
        Hair  & 0.0506 & 0.4657 & 0.642 & \textbf{0.0363} & \textbf{0.5463} & \textbf{0.725} \\
        \midrule
        Whole & 0.0471 & 0.4230 & 0.594 & \textbf{0.0432} & \textbf{0.4492} & \textbf{0.628} \\
        \bottomrule
    \end{tabular}
    }
    \label{tb:decompose}
\end{table}

We visually compare our system against the preliminary StdGEN baseline in Fig.~\ref{fig:comp_stdgen}. The baseline, constrained by its simplistic grid estimation, frequently produces topological artifacts. For instance, long skirts often exhibit severe fracturing, and delicate features like ``ahoge'' are typically lost. In contrast, StdGEN++, benefitting from its scalable system design (i.e., the dual-branch S-LRM with sparse evaluation), effectively scales to higher resolutions. This enables the direct synthesis of production-ready, coherent meshes with sharp, high-frequency details, eliminating the artifacts observed in the prototype version.

\begin{table}[h]
    \centering
    \caption{Ablation study on the Hair layer. We observe a clear stepwise improvement: high-resolution grid enhances basic details, while the facial branch further refines complex topology.}
    \resizebox{0.95\linewidth}{!}{
    \begin{tabular}{lcccc}
        \toprule
        Model Variant (Hair Layer) & CD$\downarrow$ & IoU$\uparrow$ & F1$^{0.5}$$\uparrow$ \\
        \midrule
        StdGEN (Baseline) & 0.0506 & 0.4657 & 0.6416 \\
        + Coarse-to-Fine Proposal & 0.0421 & 0.5229 & 0.6995 \\
        + Facial Branch (\textbf{Final}) & \textbf{0.0363} & \textbf{0.5463} & \textbf{0.7245} \\
        \bottomrule
    \end{tabular}
    }
    \label{tb:ablation_hair}
\end{table}

\xhdr{Ablation: Resolution and Facial Branch}
To validate the system's modular design, we conduct an ablation study focusing on the challenging \textit{Hair} geometry (Tab.~\ref{tb:ablation_hair}).
First, activating the coarse-to-fine proposal module to upscale the resolution improves the F1$^{0.5}$ score from $0.6416$ to $0.6995$, validating that high-density voxel grids are a prerequisite for recovering thin structures.
Crucially, the integration of the specialized \textit{Facial S-LRM Branch} yields the best performance, boosting the F1$^{0.5}$ score to $0.7245$. This monotonic improvement confirms that our multi-branch strategy provides essential semantic priors, enabling the reconstruction of intricate hairstyle topologies that resolution scaling alone cannot resolve.

\begin{figure}[htbp]
\centering
\includegraphics[width=1\linewidth]{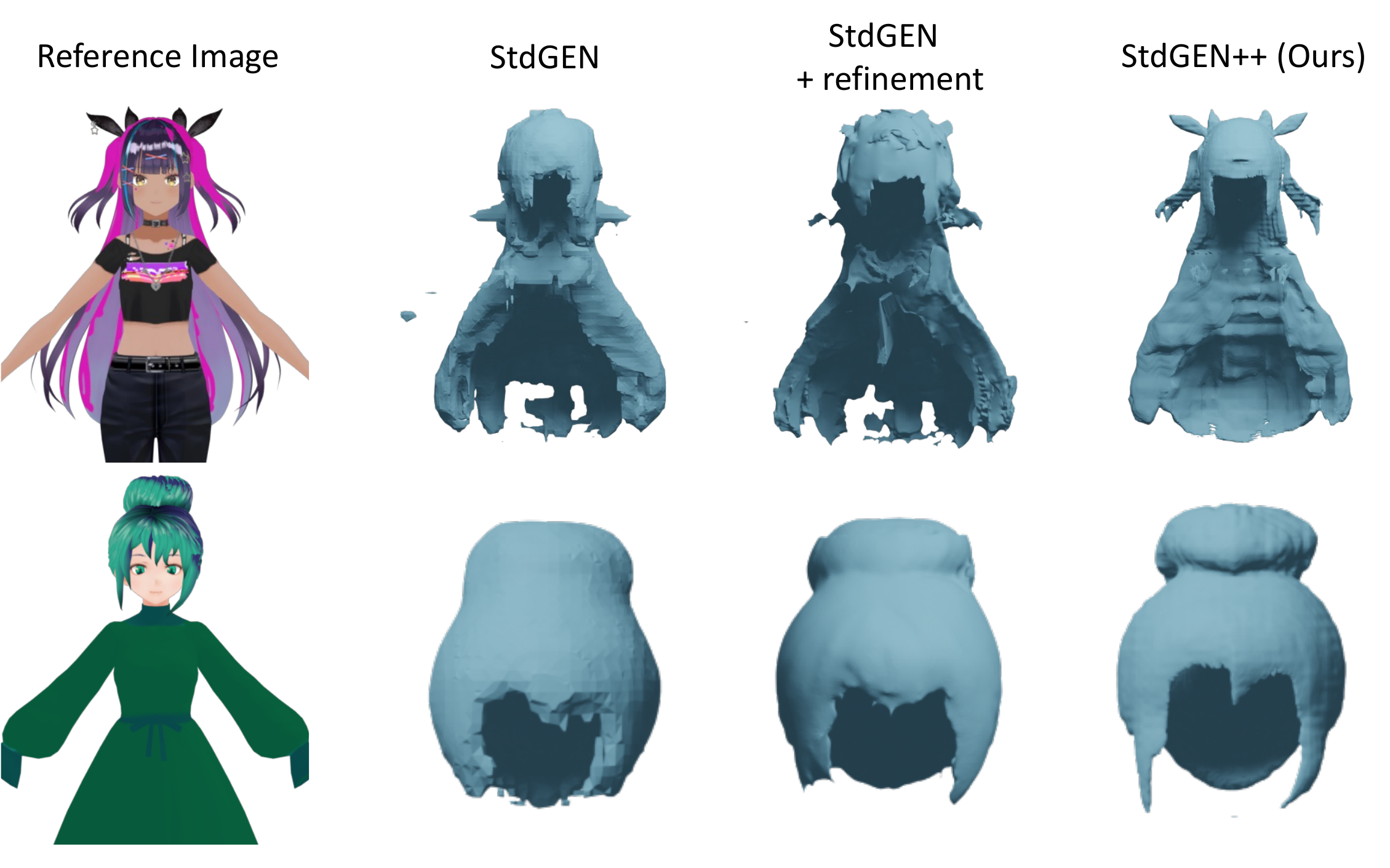}
\caption{\textbf{Limitations of test-time refinement versus high-resolution reconstruction.}
While the multi-view refinement used in StdGEN can smooth surfaces, it fails to repair fundamental topological defects like holes in complex hair structures (middle column). Our method (right column) fundamentally resolves these issues by scaling up the reconstruction resolution, yielding structurally complete geometry even without refinement.}
\label{fig:refine_limit}
\end{figure}

\xhdr{Analysis on Multi-layer Refinement}
We re-evaluate the test-time refinement strategy from the perspective of pipeline efficiency and fidelity. Fig.~\ref{fig:refine_limit} highlights a critical limitation of the post-processing paradigm: optimization-based refinement relies on a valid initial topology. As seen in the baseline results, when the base mesh contains topological defects (e.g., holes), refinement merely smooths the artifact boundaries without repairing the geometry.
In contrast, our StdGEN++ system resolves these structures correctly at the source via high-resolution inference, rendering computationally expensive post-hoc topological repair unnecessary.
Furthermore, quantitatively, we find that applying refinement to our high-fidelity outputs can be counterproductive for thin structures (e.g., slight degradation in Hair CD/visual sharpness). Consequently, to streamline the system workflow without compromising quality, we apply refinement selectively only to the body layer, while relying on the direct high-fidelity output of the S-LRM for cloth and hair.

\begin{figure}[htbp]
\centering
\includegraphics[width=1\linewidth]{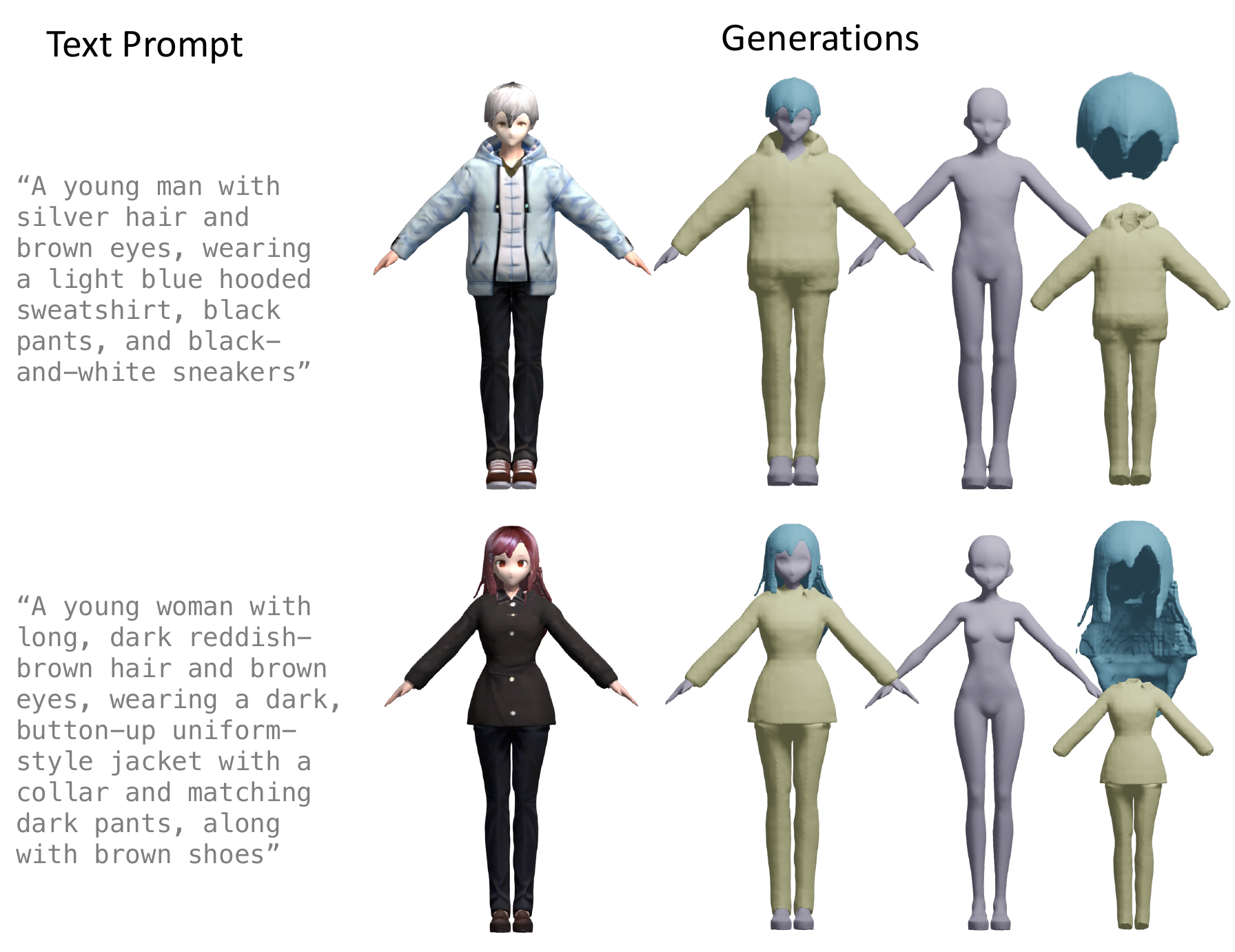}
\caption{\textbf{Text-conditioned layered character generation.}
By leveraging the unified A-pose intermediate representation, our pipeline seamlessly converts abstract text prompts into canonical visual priors, which are then processed into semantically decomposed, industrial-ready 3D meshes (Right).}
\label{fig:text_gen}
\end{figure}

\xhdr{Integrated Text-to-Character Generation}
To demonstrate the industrial compatibility of our comprehensive system, we showcase its performance under the pure text-conditioned modality (as defined in Sec.~\ref{subsec:multiview}).
In practical production pipelines, character assets often originate from high-level textual descriptions rather than finished concept art.
Our system addresses this by utilizing the canonical A-pose as a unified intermediate interface. As shown in Fig.~\ref{fig:text_gen}, we integrate a fine-tuned Diffusion module to translate natural language prompts into standardized A-pose priors. These intermediate representations are then seamlessly processed by our coarse-to-fine S-LRM to yield high-fidelity, semantically decomposed 3D meshes.
This workflow proves that our system is not limited to image-to-3D reconstruction but serves as a flexible, holistic solution capable of bridging the gap between abstract creative intent (text) and physically usable digital assets (decomposed layered meshes), significantly streamlining the character creation pipeline.

\begin{figure}[htbp]
\centering
\includegraphics[width=0.9\linewidth]{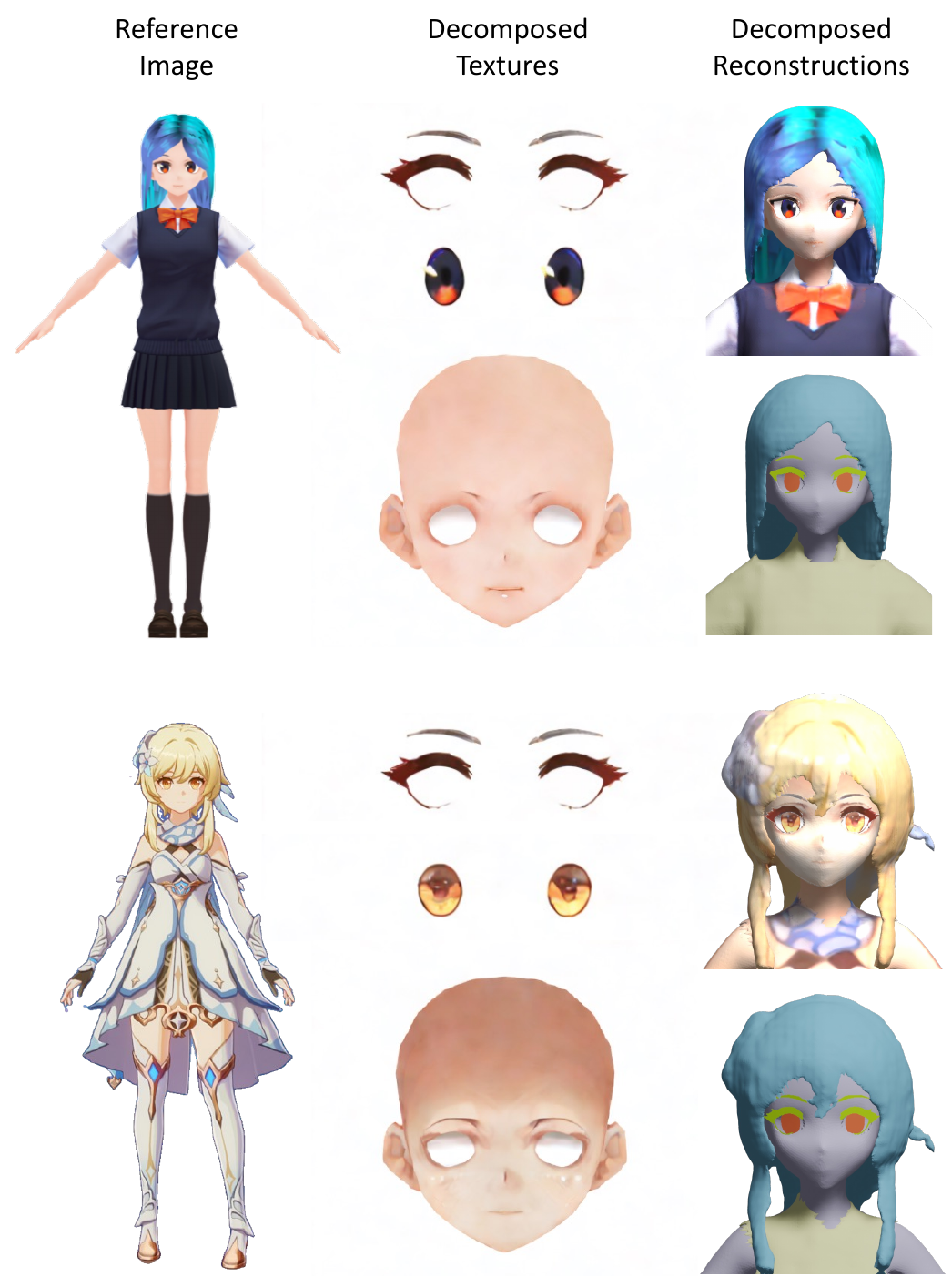}
\caption{\textbf{Semantic texture decomposition results.} 
Our system decomposes facial appearance into editable, industry-standard layers. 
\textbf{Left:} Reference image. 
\textbf{Middle:} Decomposed maps (eyebrow/eyelash, iris, and base skin) generated by our video-diffusion module. 
\textbf{Right:} Composited 3D character (top) and semantic visualization (bottom), showing precise alignment between textures and geometry. This supports independent manipulation, like gaze redirection, without distortion.}
\label{fig:tex_decomp}
\end{figure}

\subsection{Texture Decomposition and Editability}
Beyond geometric layering, our comprehensive system addresses the semantic disentanglement of appearance—a critical requirement for animation and gaming workflows. We evaluate the performance of our semantic texture decomposition module in Fig.~\ref{fig:tex_decomp}.

\xhdr{Visual Fidelity and Separation}
As illustrated in the middle column of Fig.~\ref{fig:tex_decomp}, our video-diffusion-based approach successfully isolates anatomical components into dedicated texture maps.
Unlike simple segmentation, our method generates generatively inpainted backgrounds for each layer.
Specifically, observe the \textit{base skin} layer (Middle, Bottom): the system effectively ``imagines'' and reconstructs the clean skin and white sclera areas that were originally occluded by the large anime irises and lashes. This eliminates the ``ghosting'' artifacts common in monolithic texture projection.
Simultaneously, the \textit{iris} and \textit{eyebrow} layers (Middle, Top) are extracted with sharp boundaries and high transparency precision, ensuring they can be overlaid seamlessly onto the base skin.

\xhdr{Industrial Compatibility}
The rightmost column confirms that these decomposed textures map correctly onto the generated 3D geometry. This layered representation mirrors professional layouts, directly enabling downstream tasks that were previously impossible with StdGEN's monolithic output. For example, the independence of the iris texture allows for gaze tracking (moving the iris UV without warping the skin) and appearance editing (e.g., changing eye color or eyebrow) by simply modifying the respective texture layer, validating the system's enhanced compatibility with modern pipelines.

\subsection{Applications}

\begin{figure}[htb]
\centering
\includegraphics[width=0.95\linewidth]{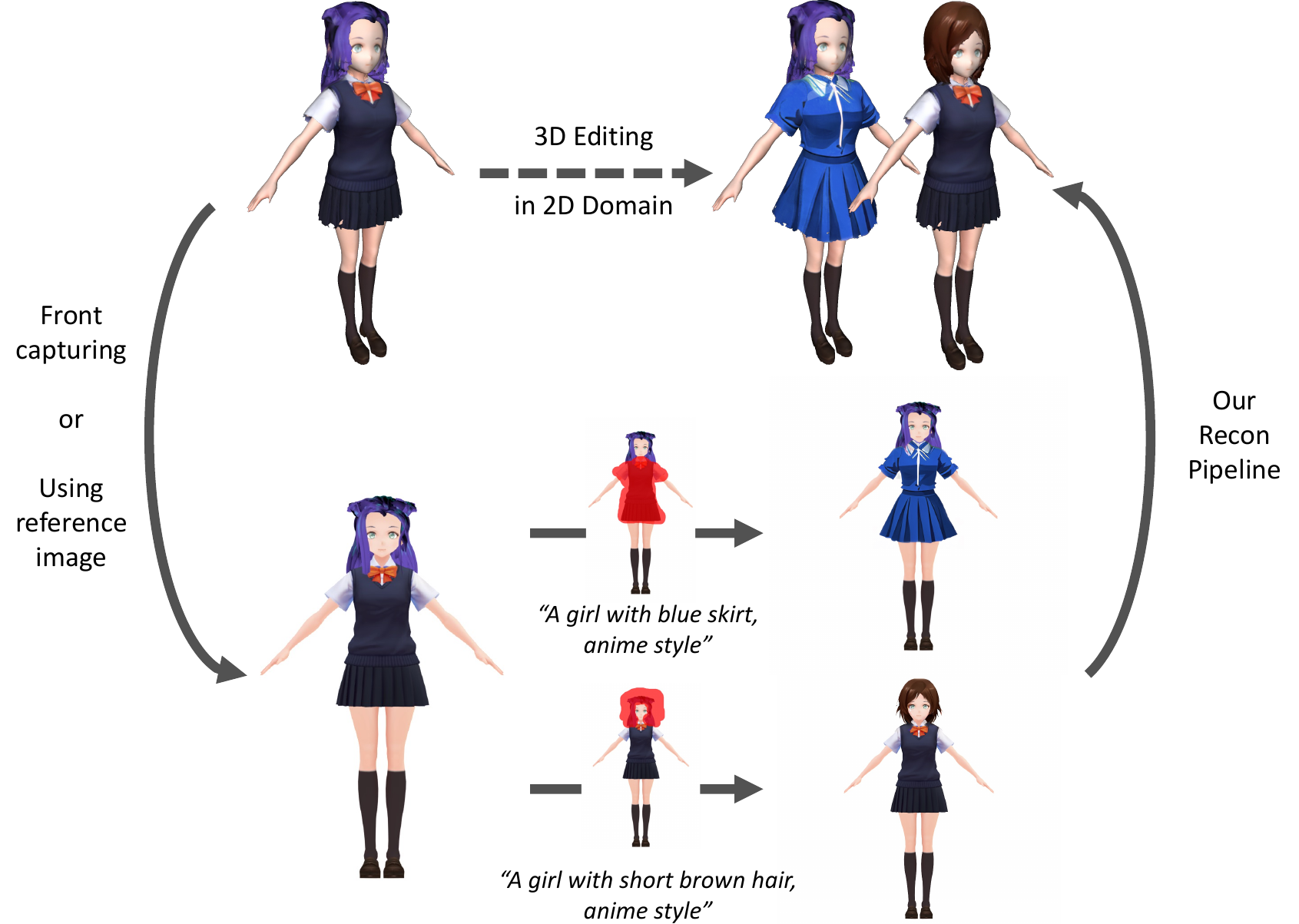}
\caption{Our pipeline enables diverse 3D editing using only text prompts, masks, and in-painting diffusion in the 2D domain.}
\label{fig:edit}
\end{figure}

\xhdr{3D Editing via 2D In-painting}
Our system's modular architecture naturally facilitates 3D editing by bridging it with mature 2D generation tools. Unlike monolithic reconstruction methods that require regenerating the entire mesh for local changes, our framework supports non-destructive, layer-wise customization.
As illustrated in Fig.~\ref{fig:edit}, users can modify specific components (e.g., outfit or hairstyle) using a streamlined workflow: starting with the generated A-pose view, a user provides a crude mask and a text prompt to an off-the-shelf in-painting model (e.g., HD-Painter~\cite{manukyan2023hd}).
Crucially, because our underlying S-LRM is semantically disentangled, the modified 2D region can be independently reconstructed into a new 3D layer and seamlessly swapped with the original component, while the remaining layers (e.g., the base body) are preserved intact. This capability significantly lowers the barrier for creating diverse 3D variations from a single reference.

\begin{figure}[t]
\centering
\includegraphics[width=0.9\linewidth]{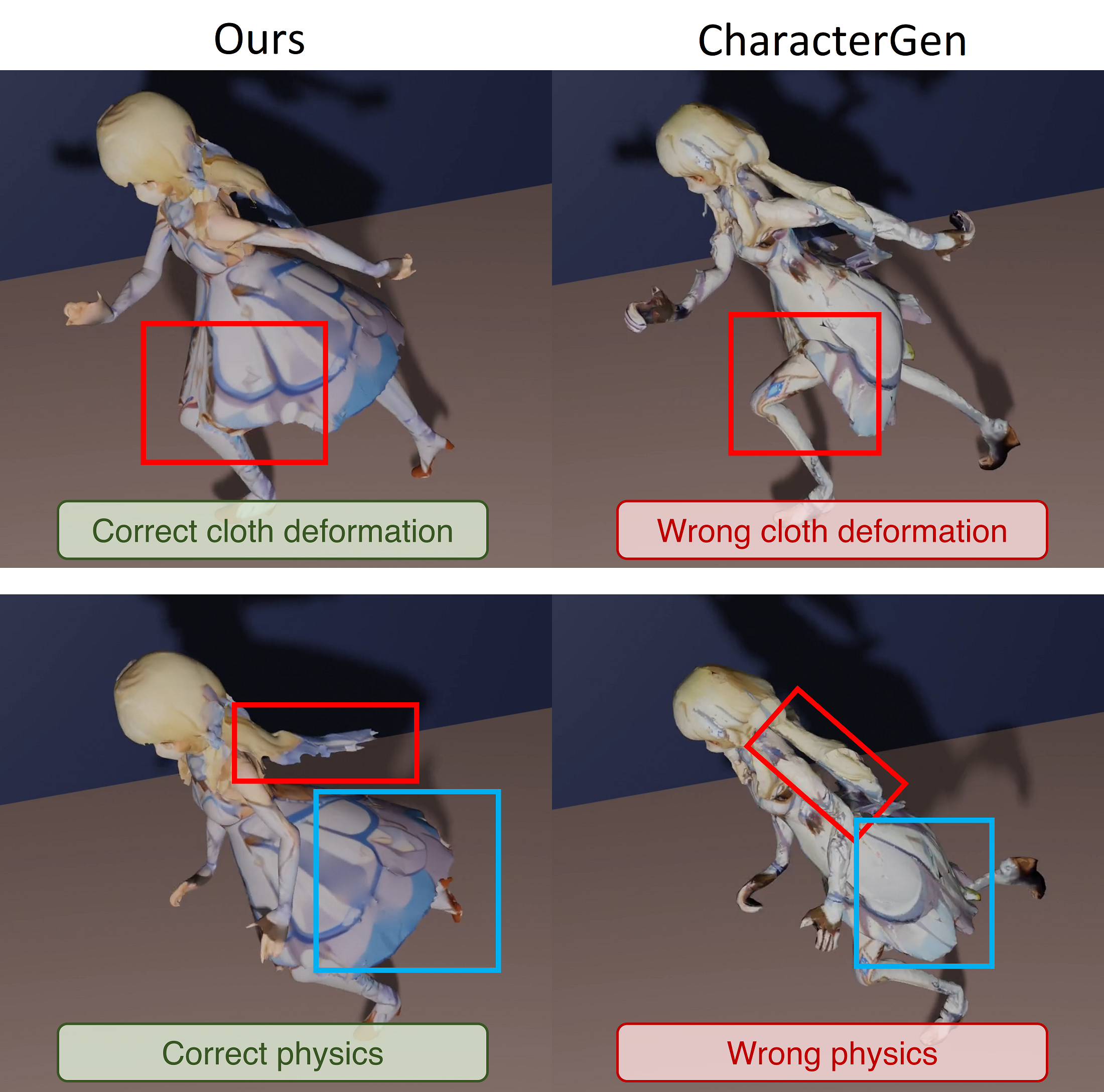}
\caption{Rigging and animation comparisons on 3D character generation. Our method demonstrates superior performance in human perception and physical characteristics.}
\label{fig:anim}
\end{figure}

\xhdr{Physics-Ready Animation}
The structural superiority of our decomposed, hollow geometry is most evident in downstream animation tasks.
We rig and animate characters generated by our method and CharacterGen~\cite{peng2024charactergen} for comparison (Fig.~\ref{fig:anim}).
Existing monolithic methods suffer from ``mesh gluin'' artifacts, where hair and clothing are topologically fused to the body skin, leading to unnatural stretching and distortion during movement.
In sharp contrast, our approach produces physically independent layers—the clothing is a standalone hollow mesh detached from the body, and the hair is separated from the face.
This independence not only prevents rigging artifacts but also enables advanced physics simulations (e.g., cloth dynamics and hair swing) that align with professional animation standards, functionality that is structurally impossible for non-decomposed baselines.

\begin{figure}[t]
\centering
\includegraphics[width=1.0\linewidth]{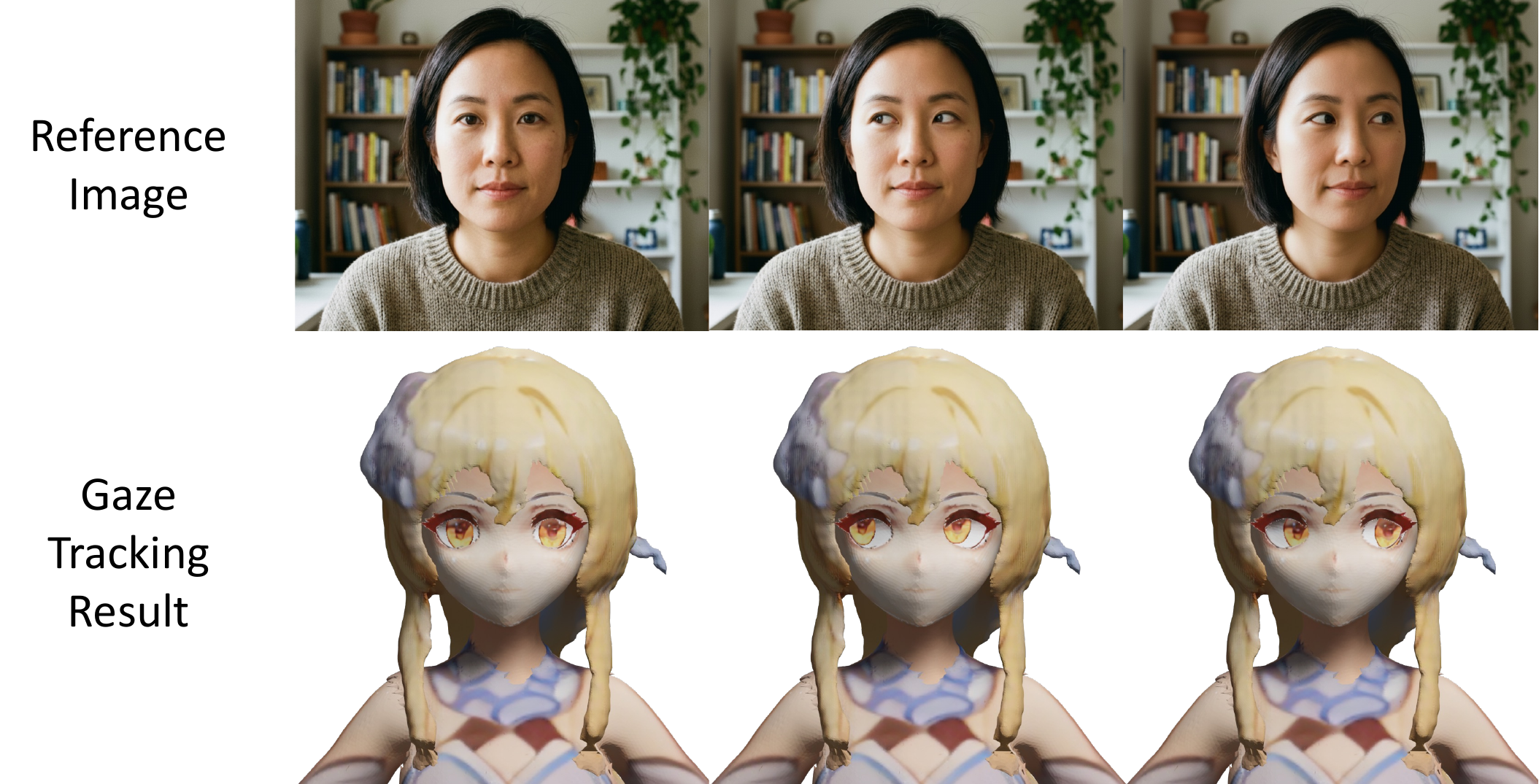}
\caption{Gaze tracking demonstration. By applying transforms to the independent iris layer, the character's gaze can be redirected to match the input.}
\label{fig:gaze}
\end{figure}

\xhdr{Gaze Tracking}
A direct benefit of our semantic texture decomposition is the enablement of gaze tracking. In traditional monolithic reconstruction, eyes are typically ``baked'' into the facial geometry, making independent movement difficult without creating texture artifacts.
In contrast, our system generates a dedicated floating iris layer and a fully inpainted clean sclera (eye white) layer.
Fig.~\ref{fig:gaze} demonstrates this structural advantage by transferring gaze directions from a reference video to the generated character.
The result shows smooth eye movement where the iris glides naturally over the sclera without revealing any "ghosting" artifacts. This demonstrates that our decomposed assets are structurally ready to be bound to facial control rigs for expressive animation tasks.

\section{Conclusion}

In this work, we present StdGEN++, a comprehensive system that unifies diverse inputs into high-fidelity, semantically decomposed 3D characters. Empowered by the Dual-branch S-LRM, efficient surface extraction schemes, and dedicated diffusion models, our framework ensures true semantic disentanglement, producing structurally independent mesh layers (e.g., hollow clothing) and editable texture components (e.g., separated iris) that align with industrial standards. Extensive experiments demonstrate that our method surpasses existing baselines in geometry, texture, and decomposability; furthermore, its structural independence unlocks advanced capabilities including non-destructive editing, physics-compliant animation, and gaze redirection, marking a significant step toward automated, production-ready character creation.

\bibliographystyle{IEEEtran}
\bibliography{main}

\end{document}